\documentclass{article}
\pdfoutput=1
\usepackage{lmodern}
\usepackage{microtype} 
\usepackage[a4paper, margin=1in]{geometry}
\usepackage{parskip}
\usepackage{setspace}  
  
\usepackage{mathrsfs}
\usepackage[utf8]{inputenc} %
\usepackage[T1]{fontenc}    %
\usepackage{hyperref}       %
\usepackage{url}            %
\usepackage{booktabs}       %
\usepackage{amsfonts}       %
\usepackage{nicefrac}       %
\usepackage{microtype}      %
\usepackage{lipsum}         %
\usepackage{graphicx}

\usepackage[numbers]{natbib}
\usepackage{bibunits}

\usepackage{doi}
\usepackage{caption} 
\usepackage{adjustbox}
\usepackage{subcaption}
\usepackage{xcolor}
\usepackage{tcolorbox}
\usepackage{tikz}
\usepackage{makecell}
\usepackage{array}
\usepackage{multirow}
\usepackage{multicol}
\usepackage{amsmath}
\usepackage{amssymb}
\usepackage{mathtools}
\usepackage{amsthm}
\usepackage{textcomp}
\usepackage{algorithm}
\usepackage{algorithmic}

\usepackage{authblk}

\usepackage[capitalize,noabbrev]{cleveref}

\usepackage{pifont}     

\usepackage[final, commandnameprefix=ifneeded]{changes}  %

\title{\bf How simple can you go? An off-the-shelf transformer approach to molecular dynamics}

\author[1,2]{Max Eissler}
\author[1,2]{Tim Korjakow}
\author[5]{Stefan Ganscha}
\author[5]{Oliver T.\ Unke}
\author[1,2,3,4,5]{Klaus-Robert M\"uller}
\author[1,2]{Stefan Gugler\thanks{Corresponding author: \texttt{stefan.gugler@tu-berlin.de}}\hspace{0.05cm}}

\affil[1]{BIFOLD – Berlin Institute for the Foundations of Learning and Data}
\affil[2]{Machine Learning Group, Technische Universit\"at Berlin}
\affil[3]{Department of Artificial Intelligence, Korea University}
\affil[4]{Max-Planck Institute for Informatics}
\affil[5]{Google DeepMind}

\date{Jan 28, 2026}

\begin{document}
\begin{bibunit}
\maketitle

\begin{abstract}
Most current neural networks for molecular dynamics (MD) include physical inductive biases, resulting in specialized and complex architectures. This is in contrast to most other machine learning domains, where specialist approaches are increasingly replaced by general-purpose architectures trained on vast datasets. In line with this trend, several recent studies have questioned the necessity of architectural features commonly found in MD models, such as built-in rotational equivariance or energy conservation. In this work, we contribute to the ongoing discussion by evaluating the performance of an MD model with \textit{as few specialized architectural features as possible}. We present a recipe for MD using an edge transformer (ET), an ``off-the-shelf'' transformer architecture that has been minimally modified for the MD domain, termed MD-ET. Our model implements neither built-in equivariance nor energy conservation. We use a simple supervised pretraining scheme on $\sim$30 million molecular structures from the QCML database. Using this ``off-the-shelf'' approach, we show state-of-the-art results on several benchmarks after fine-tuning for a small number of steps. Using MD-ET as a simple and expressive testbed, we examine the effects of being only approximately equivariant and energy conserving for MD simulations and thereby try to evaluate the practical usefulness of unconstrained MD models. While our model exhibits runaway energy increases on larger structures, we show approximately energy-conserving NVE simulations for a range of small structures.
\end{abstract}

\section{Introduction}
\label{sec:intro}

Molecular dynamics (MD) simulations are essential for understanding molecular behavior \citep{frenkel2023,tuckerman2023,marx2009}, such as molecular relaxation \citep{schlegel2011}, predicting structure \citep{geng2019}, modeling interactions \citep{chakraborty2021}, replicating spectra \citep{ditler2022}, up to protein folding and design \citep{jumper2021,unke2024, abramson_accurate_2024}. Machine Learning (ML) models can be used as heuristic approximations of the Schr\"{o}dinger equation thereby significantly accelerating accurate MD simulations (e.g.~\cite{rupp2012fast,noe2020machine,keith2021combining,unke2021machine}). To ensure physical consistency of predictions, researchers have focused on incorporating inductive biases into MD models, which restrict models to physically plausible solutions \citep{chmiela2017,schutt2017schnet,schutt2018, thomas2018,chmiela2018towards,unke2019physnet}. For instance, rotational equivariance guarantees that if a molecule is rotated, the forces are rotated as well (e.g.~\cite{schutt2021equivariant,bronstein2021,batzner2022nequip}).
In contrast to this, many other ML domains are increasingly using unconstrained, general-purpose architectures and large amounts of training data \citep{kaplan_scaling_2020, esser_scaling_2024, dosovitskiy_image_2020}.

Several recent theoretical \citep{gerken2024,nordenfors2024} and empirical works \citep{bigi2024, bigi2026_limits_unconstrained_mlip, ouderaa2022, elhag2024, brehmer2024} discuss the necessity of commonly used physical constraints and inductive biases in MD models, most notably SO(3)-equivariance and energy conservation. 
For example, architectures predicting non-conservative forces include ForceNet \citep{hu2021}, GemNet-dT \citep{gasteiger2021}, Orb \citep{neumann_orb_2024, rhodes2025_orbv3} or Equiformer \citep{liao2024}.

\added{
Physical inductive biases dramatically simplify the problem of learning molecular force fields by reducing the dimensionality of the functional solution space and are thus easy to motivate. There are also, however, several motivations to remove them:
Common ways to implement equivariance such as group convolutions or irreducible representations are typically computationally more expensive and drastically restrict the design of the neural architecture \cite{duval2023hitchhiker}. 
Furthermore, there is some theoretical evidence that equivariant networks have inferior training dynamics \cite{nordenfors2024}. 
Meanwhile deriving energy-conserving forces using automatic differentiation increases wall time by 2--3x, more when considering that automatic optimization frameworks such as \texttt{torch.compile} do not natively support second derivatives as required to train conservative force fields \cite{pytorch_issue_91469}, though framework-specific workarounds exist \cite{tan2025high}. Furthermore, the need to compute second derivatives only allows the use of components that are smooth everywhere, which, for instance, some activation functions are not.
Finally, there are several ML subfields such as graph learning and ML on point clouds which are concerned with problems closely related to MD. If physical inductive biases could be relaxed or approximated sufficiently closely by a general-purpose model these fields could be unified \cite{pozdnyakov2024}.
}

We thus study an extreme case: \textit{We present and evaluate the best MD model we can build using a minimal amount of MD-specific design features.} Specifically, we aim to use an architecture that could be used in some related ML setting with minimal changes. \added{Our aim is to explore potential benefits of a fully unconstrained approach to MD, while highlighting problems arising from a lack of physical inductive biases.}

Our ``off-the-shelf'' recipe is simple: We modify an edge transformer (ET) \citep{bergen2021, muller2024} with MD-specific embedding layers and train on the new QCML database~\cite{ganscha2025qcml}. QCML ranges across the periodic table, including out-of-equilibrium structures, different spin and charge states, and includes properties for a subset of $\sim$30 million entries calculated with density functional theory accuracy at the PBE0 level \citep{perdew1996,adamo1999} (including dispersion corrections \cite{caldeweyher2019generally, hermann2020density}). 
During training, we use two randomly rotated and mirrored copies of each sampled structure, so that our model can learn approximate O(3)-equivariance. Instead of predicting a combined loss \citep{mailoa2019} of energies and forces or calculating forces as the negative gradient of the predicted energy with respect to positions, we predict forces directly. \deleted{This approach does not guarantee that the predicted forces are energy-conserving, but it improves the speed of the model, because forces need not be calculated by automatic differentiation.} For downstream task evaluation, we fine-tune the model for a small number of steps only. 

For MD simulations, we evaluate either the pretrained model without further training (zero-shot) or fine-tune using a small number of samples of the target structure (few-shot).
Our contributions are as follows:

\begin{enumerate}
    \item We present an MD-adapted edge transformer (MD-ET) and show that our implementation can achieve competitive performance on several common benchmarks. We also demonstrate sample-efficient fine-tuning. 
    \item We examine the effects of forgoing almost all commonly used inductive biases. Specifically, we evaluate the effects of being only approximately equivariant and energy-conserving in detail. 
    \item We present a comparison of MD-ET with an identically pretrained equivariant and energy-conserving model as well as the recently released Orb-v2 model, highlighting the impact of removing physical constraints on MD simulations.
\end{enumerate}

\section{Background}
\label{sec:rw}
While incorporating physics-inspired inductive biases into machine learning force fields (MLFF) models yields powerful and data-efficient architectures, they also impose constraints. For instance, directly predicting (non-conservative) rotationally equivariant forces restricts a model to only performing equivariant operations \cite{duval2023hitchhiker}. A more subtle downside of specialized architectures is the increased difficulty to transfer advances from other ML research and profit from improved hardware and software tools. Engineering complexity was recently cited as the reason the AlphaFold project removed its equivariant inductive biases \citep{abramson_accurate_2024, jumper2021}. Since protein conformer prediction is thematically close to MD, this prompted discussions on which inductive biases are essential to the MD domain: Questioning the need for equivariance, \citet{langer2024} find that data augmentation during training can achieve a high degree of approximate equivariance in MD. \citet{neumann_orb_2024} present Orb, a completely unconstrained message-passing architecture with outstanding benchmark performance \added{and it has also been used in combination with quantum-mechanical analysis for orbital localization in polymer nanocomposites} \citep{farmer2025acsomega_orb_application}. \citet{elhag2024} show that an additional loss term to induce equivariance combined with data augmentation results in competitive performance and approximate equivariance. \citet{wang2024} demonstrate the potential benefits of unconstrained approaches in molecular conformer generation, achieving state-of-the-art results without relying on domain-specific assumptions. \added{To demonstrate that methods designed to process point clouds can be used for MD, \citet{pozdnyakov2024} introduce PET, a message-passing architecture where each message-passing block uses a transformer to process local neighborhoods. Specifically, for each atom-centered environment, tokens encoding neighbor information (positions, species, incoming messages) undergo self-attention, with output tokens serving as outgoing messages. While the locality of neighborhoods gives PET linear runtime, self-attention over all neighbor tokens is costly since the same atom pairs contribute to multiple overlapping neighborhoods. PET-based potentials have already been used in materials simulations, e.g., to study reconstructed the $\beta$-lithium thiophosphate surface, $\beta$-Li$_3$PS$_4$ \citep{turk2025lithium_thiophosphate_surfaces}. Furthermore, PET-MAD \citep{mazitov2025petmad} was developed as a generally applicable interatomic potential trained on a dataset designed to increase atomic diversity \citep{mazitov2025mad}.}
\citet{qu2024} propose a scalable attention-based MLFF, EScAIP, which improves computational efficiency and model expressivity by leveraging attention mechanisms and optimized GPU kernels, achieving state-of-the-art performance across various chemical datasets. \added{The same group has proposed a graph-less model, exploring even further reduction of inductive bias \cite{kreiman2025transformers_graphless}.}{} Several theoretical works also suggest that unconstrained models can produce equivariant outputs under certain circumstances \cite{gerken2024, nordenfors2024, puny_frame_2021}. In contrast, the performance effects of removing energy conservation are less clear. Although benchmarking results suggest non-conservative models can perform stable MD simulations \cite{neumann_orb_2024}, \citet{bigi2024} report that all non-conservative models in their study suffer from runaway temperature increases and thus instability. 

Current debates in the field of ML for MD center around competing paradigms---one leveraging inductive biases for sample efficiency, and another prioritizing simplicity, scale and transfer using general-purpose models. Recently, several aspects of the scaling paradigm have been studied and discussed in isolation \cite{langer2024, bigi2024, brehmer2024, wood2025_uma, fu2025}. Our work advances this discourse by empirically evaluating a simple baseline approach that fully adopts the scaling paradigm for MD and exploring its limits.

\section{Molecular dynamics edge transformer}
\label{sec:mdet}

We use the transformer, an architecture which has proven to scale well in many domains, as a starting point in our design process. 
\added{Since we want to keep the architecture as close as possible to a regular transformer implementation we also choose to forgo graph construction and thus cutoffs completely. Note that any attention mechanism can be converted into an equivalent sparse message passing algorithm by only evaluating the dot products selectively. Any attention mechanism converted in such a manner will also exhibit a linear runtime. We thus do not see a higher than linear-scaling runtime as an issue for this analysis, even though it is prohibitive for many typical use cases of MLFFs.
}

\added{
We choose to represent an atomic system using its own frame of reference, i.e., using pairwise distances. While this represents a significant inductive bias, it removes several symmetries from the representation without constraining the model architecture. To enable the model to learn equivariant functions we also supply pairwise directional embeddings using absolute polar and azimuthal angles.
}

\added{Expressivity, as employed in the context of graph learning, is the ability of an ML model to differentiate two non-equivalent inputs, i.e., to distinguish two graphs that are not isomorphic. \citet{morris2023weisfeiler} show that the expressivity of a graph learning model can be related to performing a variant of the Weisfeiler--Leman (WL) graph isomorphism test. Additionally more expressive models can learn more complex features in each layer, which implies that even if a less expressive model can still theoretically learn to distinguish two inputs it might have more difficulty converging to a representation which separates them given the same number of layers \cite{muller2024}. 
}

\added{
\citet{hordan2024weisfeiler} show that using pairwise distance features alone a 3-WL neural network can distinguish any two euclidean point clouds. Since we use pairwise distance embeddings we thus believe that 3-WL expressivity is a feature for which it is worthwhile to deviate from the simplest possible transformer implementation. The Edge Transformer (ET) variant introduced by \citet{bergen2021} uses a triangular self-attention implementation which is similar to standard dot-product self-attention (see Appendix \ref{sec:et-impl-details}) and has been shown to be 3-WL expressive \cite{muller2024}. Since it uses a mixture of dot-product attention and sum-pooling it also features a runtime and memory complexity of $\mathcal{O}(N^3)$, where $N$ is the number of atoms, whereas full edge-to-edge self-attention has a complexity of $\mathcal{O}(E^2) = \mathcal{O}(N^4)$ without cutoffs. Merely using atom-to-atom self-attention could further reduce complexity to $\mathcal{O}(N^2)$, atom-wise tokenization however only allows the use of edge features as (learned) biases in the attention operation, a restriction which we find too limiting. We thus choose to use ET \cite{bergen2021} for this study. As noted above we use the "off-the-shelf" dense implementation of ET, which is sufficiently fast for molecules smaller than approximately 100 atoms (see Appendix \ref{si:timings}) and merely add embedding layers for MD. We therefore name our variant MD-ET.
}

\deleted{
We decide to use the ET variant introduced by \mbox{\citet{bergen2021}} for algorithmic reasoning and adapted by \mbox{\citet{muller2024}} for graph learning due to its expressivity (see below). ET is more expressive than a regular transformer due to its higher-order attention mechanism. We modify only the ET's embeddings for MD and therefore name our variant MD-ET.
}

\paragraph{Expressivity} \deleted{Expressivity describes the ability of an ML model to differentiate two non-equivalent inputs, i.e., to distinguish two graphs that are not isomorphic. 
\mbox{\citet{morris2023weisfeiler}} show that the expressivity of a graph learning model can be related to performing a variant of the Weisfeiler--Leman (WL) graph isomorphism test. 
\mbox{\citet{hordan2024weisfeiler}} extend the WL-test to equivariant point clouds and show that a cutoff-free model with 3-WL expressivity is universal on 3D point clouds, i.e., such a model can theoretically learn to distinguish any two 3D point clouds. \mbox{\citet{muller2024}} show the ET has an expressive power equivalent to 3-WL. The ET can thus universally distinguish 3D point clouds, making it theoretically well-suited as an MLFF. 
}
\paragraph{Tokenization} The ET derives its name from its tokenization scheme. We represent a molecular system as the set of all edges of a fully connected graph, that is, all pairs of atoms (including self-loops). We then encode the molecular system as a three-dimensional tensor $\boldsymbol{x} \in \mathbb{R}^{N\times N\times D}$, where $N$ denotes the number of atoms in the system and $D$ is the embedding dimension. An embedding $\boldsymbol{x}_{ij}$ thus represents the edge between atoms $i$ and $j$, while $\boldsymbol{x}_{ii}$ represents the atom $i$.

\paragraph{Triangular attention mechanism} To achieve the expressive power of 3-WL, the ET performs updates on $\boldsymbol{x}$ using a triangular attention mechanism, performing interactions between triples of atoms, 
\begin{equation}
    \operatorname{TRIA}(\boldsymbol{x}_{ij}) = \sum_{l=1}^N \alpha_{ilj}\boldsymbol{v}_{ilj} \ ,
\end{equation}
a tensor product between an attention tensor $\boldsymbol{\alpha} \in \mathbb{R}^{N \times N \times N}$ and a value tensor $\boldsymbol{v} \in \mathbb{R}^{N \times N \times D}$. An element of the attention tensor is calculated as 
\begin{equation}
    \alpha_{ilj} = \operatorname{softmax}_{l\in[N]}
    \left(
        \frac{1}{\sqrt{d}}
        \boldsymbol{x}_{il}
        \boldsymbol{W}^Q
        \left(\boldsymbol{x}_{lj}\boldsymbol{W}^K\right)^\top
    \right) \in \mathbb{R} \ ,
\end{equation}
i.e., the attention score between the representations of edges \textit{(i, l)} and \textit{(l, j)} and where the notation $[N]$ represents the set $\{1, 2, \ldots, N\}$. 
The value vector 
\begin{equation}
    \boldsymbol{v}_{ilj} = \boldsymbol{x}_{il}\boldsymbol{W}^{V_1} \odot \boldsymbol{x}_{lj}\boldsymbol{W}^{V_2} \in \mathbb{R}^{D}
\end{equation}
is a combination of the value vectors of $\boldsymbol{x}_{il}$ and $\boldsymbol{x}_{lj}$ through elementwise multiplication $\odot$. $\boldsymbol{W}^Q$,$\boldsymbol{W}^K$,$\boldsymbol{W}^{V_1}$, $\boldsymbol{W}^{V_2}$ $\in \mathbb{R}^{D\times D}$ are learned weight matrices \cite{muller2024}. The triangular attention mechanism extends to multi-head attention analogous to regular self-attention. Similarly, a full ET layer is defined analogously to a transformer layer \cite{vaswani_attention_2017}; specifically, 
  \begin{equation}
\boldsymbol{x}^{(t)}_{ij}
=\operatorname{FFN}\left(
\operatorname{TRIA}\left(
\operatorname{LN}\left(
\boldsymbol{x}^{(t-1)}_{ij}
\right)
\right)
+ \boldsymbol{x}^{(t-1)}_{ij}
\right),
\end{equation}
where $\operatorname{FFN}$ is a feed-forward neural network and $\operatorname{LN}$ denotes layer norm \cite{lei2016layer}.

\paragraph{Molecular embeddings}
To generate initial edge representations, we combine several embeddings: i) spin and charge, ii) atomic numbers, iii) pairwise distances, and iv) pairwise displacement vectors. To maintain MD-ET's simplicity, we implement commonly used embedding layers where possible. For details, see Appendix \ref{sec:embedding_layer_details}.

\paragraph{Limitations}
\deleted{
ET's triangular attention is expressive and straightforward to implement and JIT-compile (see Appendix \ref{sec:et-impl-details}). The large 3D tensor products are embarrassingly parallelizable and thus execute efficiently on accelerators. However, this dense formulation of MD-ET has a runtime and memory complexity of $\mathcal{O}(N^3)$, where $N$ is the number of atoms in the system (see Appendix \ref{si:timings}). The presented version of MD-ET is thus, of course, impractical to use as an MLFF for many real-world MD applications. While a sparse implementation is fairly simple and linear-scaling, it is not ``off-the-shelf'', which is why we use the dense implementation for this analysis. We thus see our MD-ET implementation less as a general-purpose MD model and more as a simple and expressive test platform to evaluate the strengths and shortcomings of unconstrained models.
}

\section{Experiments}
\label{sec:exp}

To contribute to the current discussion of inductive biases in MD, we need to thoroughly evaluate MD-ET. Specifically, we want to address the following questions:

\begin{itemize}
    \item Can an ``off-the-shelf'' model without inductive biases compete on common benchmarks measuring accuracy, inference speed, and MD stability?
    \item How equivariant is MD-ET? Are there systematic biases depending on system orientation? \deleted{How do these biases affect the stability of the NVE simulation?}
    \item Is MD-ET approximately energy-conserving? How stable are NVE simulations?
    \item How faithful are NVT simulations? Are observables impacted by non-conservative/non-equivariant force predictions?
\end{itemize}

\label{sec:pretraining}
We use a pretrained model for all downstream evaluations.
However, instead of the commonly used denoising task \citep{zaidi_pre-training_2022, neumann_orb_2024}, we perform supervised pretraining on the new QCML dataset \citep{ganscha2025qcml}.
We create an approximate 90\%/5\%/5\% split from QCML.
As QCML contains multiple conformations for each structure sampled along their normal modes, we ensure that all conformations of a structure get assigned to the same split. To learn approximate equivariance, we use data augmentation:
During pretraining we duplicate each batch once and apply random rotations and reflections to both copies to form an augmented batch. We train MD-ET for 880k steps on QCML using a batch size of 1024 (i.e., 512 before data augmentation), which takes approximately 16 A100 GPU-days (see Appendix \ref{si:training_protocol}).
\label{sec:postprocessing}
Our model does not require any modifications for downstream fine-tuning tasks. For MD-simulations we adopt the net force and torque removal algorithm described in \cite{neumann_orb_2024}.
We find that removing net forces and torque from model predictions improves the stability of MD simulations while only increasing the time of a model evaluation by 10-20 percent (see Appendix \ref{si:timings}). We use net force and torque removal in all simulation experiments. 

\subsection{Unconstrained models are competitive on common MD benchmarks}
\label{sec:q1}

We evaluate the performance of our proposed MD-ET model across several standard MD benchmarks. We compare MD-ET with established MLFFs, focusing on their accuracy, stability, and efficiency. We evaluated MD-ET on QCML \citep{ganscha2025qcml}, MD17 \citep{chmiela2017}, Ko2020 \citep{ko2021}, xxMD \citep{pengmei2024} and SPICE \citep{eastman2023spice}, each presenting distinct challenges.

\subsubsection*{Pretraining performance}

\begin{table}[ht]
    \centering
    \caption{Force prediction test MAE on the QCML dataset in meV \AA$^{-1}$~for different models. }
    \begin{tabular}{lc}
        \toprule
        Model & Test MAE \\
        \midrule
        MD-ET & 29.9 \\
        SpookyNet$^\dagger$ & 32.1 \\
        PaiNN & 42.5 \\ %
        \bottomrule
    \end{tabular}
    \\\vspace{1ex}
    \small{$^\dagger$Does not separate conformers between training and testing.}
    \label{tab:qcml_results}
\end{table}

In Table~\ref{tab:qcml_results}, we present the mean absolute error (MAE) of force predictions for MD-ET, SpookyNet~\citep{unke2021spookynet}, and PaiNN~\citep{schutt2021equivariant} on the QCML evaluation set at the end of pretraining.

MD-ET achieves the lowest MAE among the compared models, outperforming both SpookyNet and PaiNN. This improvement over equivariant models like PaiNN and SpookyNet suggests that approximate equivariance, as utilized in MD-ET, does not hinder performance on such benchmarks. However, a low test error does not necessarily imply a low MD simulation error or a stable trajectory, as we will demonstrate below.

\subsubsection*{MD17-10k}
\label{sec:md17-molecule-specific-models}
\begin{table*}[t]
    \caption{Performance on the MD17-10k benchmark, comparing stability (ps) with a maximum at 300 ps, force MAE (meV \AA$^{-1}$), the distribution of interatomic distances, $h(r)$, and FPS. Higher is better is denoted $(\uparrow)$ and vice versa. Best average results in bold. Results marked with a $^*$ indicate that the molecule was included in the pretrain train split. FPS benchmarks conducted on a V100 GPU, except where marked $^{**}$, there conducted on an A100 GPU.
    }
    \footnotesize
\centering
\setlength{\tabcolsep}{2.5pt} %
\resizebox{\linewidth}{!}{%
\begin{tabular}{llccccccccc|cc}
\toprule
    Metric & Molecule & \makecell[c]{\textbf{DeepPot-SE} \\[-0.7ex] \scriptsize{\citep{zhang2018deep}}} & 
\makecell[c]{\textbf{SchNet} \\[-0.7ex] \scriptsize{\citep{schutt2017schnet}}} & 
\makecell[c]{\textbf{DimeNet} \\[-0.7ex] \scriptsize{\citep{gasteiger2020}}} & 
\makecell[c]{\textbf{PaiNN} \\[-0.7ex] \scriptsize{\citep{schutt2021equivariant}}} & 
\makecell[c]{\textbf{SphereNet} \\[-0.7ex] \scriptsize{\citep{liu2021}}} & 
\makecell[c]{\textbf{ForceNet} \\[-0.7ex] \scriptsize{\citep{hu2021}}} & 
\makecell[c]{\textbf{GemNet-dT} \\[-0.7ex] \scriptsize{\citep{gasteiger2021}}} & 
\makecell[c]{\textbf{NequIP} \\[-0.7ex] \scriptsize{\citep{batzner2022nequip}}} & 
\makecell[c]{\textbf{MD-ET} \\[-0.7ex] \scriptsize{(trained directly)} \\[-0.7ex] \scriptsize{(ours)}} &
\makecell[c]{\textbf{Orb}\\[-0.7ex] \scriptsize{\citep{neumann_orb_2024}} \\ (finetuned)} & 
\makecell[c]{\textbf{MD-ET} \\[-0.7ex] \scriptsize{(finetuned)} \\[-0.7ex] \scriptsize{(ours)}} \\
\midrule
     \multirow{5}{*}{\rotatebox[origin=c]{90}{Stability ($\uparrow$)}} %
     & Aspirin      & 9     & 26    & 54    & 159   & 141   & 182   & 192   & 300  & 124 & 300 & 300 \\
     & Ethanol      & 300   & 247   & 26    & 86    & 33    & 300   & 300   & 300 & 300 & 300  & 300 \\
     & Naphthalene  & 246   & 18    & 85    & 300   & 6     & 300   & 25    & 300 & 300 & 300  & 300 \\
     & Salicylic Acid & 300 & 300   & 73    & 281   & 36    & 1     & 94    & 300  & 256 & 300 & 300 \\
\cmidrule{2-13}
    & Average  & 213.8& 147.8& 60.0  &206.5  & 54    &195.8 &152.8 & \textbf{300} & 245 & \textbf{300} & \textbf{300} \\
\midrule
    \multirow{5}{*}{\rotatebox[origin=c]{90}{MAE ($\downarrow$)}} & Aspirin         & 21.0    & 35.6  & 10.0    & 9.2   & 3.4   & 22.1  & 5.1   & 2.3 & 4.2 & 2.4 & 4.2 \\
               & Ethanol         & 8.9   & 16.8  & 4.2   & 5.0     & 1.7   & 14.9  & 1.7   & 1.7  & 1.3 & 2.5  & \hspace{0.1cm}$1.0^*$ \\
               & Naphthalene     & 13.4  & 22.5  & 5.7   & 3.8   & 1.5   & 9.9   & 1.9   & 1.1 & 4.0 & 1.0  & 2.3 \\
               & Salicylic Acid  & 14.9  & 26.3  & 9.6   & 6.5   & 2.6   & 12.8  & 4.0     & 1.6 & 3.3 & 1.8  & 2.8 \\
\cmidrule{2-13}
               & Average             & 14.6 & 25.3  & 7.4 & 6.1 & 2.3   & 15.0 & 3.2 & \textbf{1.6} & 3.5 & 1.7 & 2.6 \\
\midrule
    \multirow{5}{*}{\rotatebox[origin=c]{90}{h(r) ($\downarrow$)}} & Aspirin       & 0.65  & 0.36 & 0.04 & 0.04 & 0.03  & 0.56 & 0.04  & 0.02 & 0.03 & 0.03 & 0.02 \\
               & Ethanol         & 0.09  & 0.21 & 0.15 & 0.15  & 0.13  & 0.86 & 0.09  & 0.08 &  0.03& 0.09 & 0.05 \\
               & Naphthalene     & 0.11  & 0.09 & 0.10 & 0.13  & 0.14  & 1.02 & 0.12  & 0.12 & 0.15 & 0.12 & 0.07 \\
               & Salicylic Acid  & 0.03  & 0.03 & 0.06 & 0.03  & 0.06  & 0.35 & 0.07  & 0.03 &  0.3 & 0.03 & 0.03 \\
\cmidrule{2-13}
               & Average         & 0.22  & 0.17 & 0.088 & 0.088 & 0.090 & 0.70 & 0.080 & 0.063 & 0.125 & 0.068 & \textbf{0.04} \\
\midrule
    \multirow{5}{*}{\rotatebox[origin=c]{90}{FPS ($\uparrow$)}} & Aspirin         & 88.0    & 108.9 & 20.6    & 85.8  & 17.5   & 137.3  & 56.8 & 8.4 & 159.8 & 68.6** & 159.8 \\
               & Ethanol         & 101.0   & 112.6  & 21.4   & 87.3     & 30.5   & 141.1  & 54.3   & 8.9 & 148.1  & 69.2** & 148.1 \\
               & Naphthalene     & 109.3  & 110.9  & 19.1   & 92.8   & 18.3   & 140.2   & 53.5   & 8.2 & 150.2 & 68.5** & 150.2 \\
               & Salicylic Acid  & 94.6  & 111.7  & 19.4   & 90.5   & 21.4   & 143.2   & 52.4   & 8.4 & 147.3  & 69.4** & 147.3 \\
\cmidrule{2-13}
               & Average             & 98.2 & 111.0  & 20.1 & 89.1 & 21.9   & 140.5 & 54.3 & 8.5 & \textbf{151.4} &   68.9**&\textbf{151.4} \\
\bottomrule
\end{tabular}
}

    \label{table:md17}
\end{table*}

The MD17 dataset~\citep{chmiela2017} consists of MD trajectories for small organic molecules and has been proposed as a benchmark by \citet{fu2022forces}. Models are evaluated not only for their prediction accuracy (MAE) but also for their simulation stability and faithfulness. Stability is measured by the time until a simulation becomes unstable (capped at 300 ps), and faithfulness is measured by $h(r)$, the distribution of interatomic distances \cite{fu2022forces}. Inference speed, which is critical for MD simulations, is also compared (see Appendix \ref{si:experimental_setup_fine-tuning} for evaluation details) in frames per second (FPS).

Table~\ref{table:md17} compares MD-ET with several state-of-the-art models. In addition to fine-tuning ET-MD for 2000 steps, which we see as the default way to use a large-scale pretrained model, we also report results for training the model directly on MD17. Since there is some overlap with the pretraining set, fine-tuning results for ethanol (marked with a *), have only limited significance. We also include fine-tuning results for Orb-v2 \citep{neumann_orb_2024}, which was released recently and is a popular unconstrained model. 

\begin{table*}[t]
    \caption{MAE of atomic forces (meV \AA$^{-1}$) for MD17 molecules recalculated with CCSD(T). Best results in bold. Molecules with a $^*$ are included in the pretrain dataset of MD-ET.}
    \hspace*{-0.9cm}

\setlength{\tabcolsep}{2.5pt} %
\resizebox{\linewidth}{!}{%

\begin{tabular}{lccccc|cc}
    \toprule
    \textbf{Molecule} 
    & \makecell[c]{\textbf{sGDML} \\[-0.7ex] \scriptsize{\citep{chmiela2018towards}}}
    & \makecell[c]{\textbf{GemNet-Q} \\[-0.7ex] \scriptsize{\citep{gasteiger2021}}}
    & \makecell[c]{\textbf{GemNet-T} \\[-0.7ex] \scriptsize{\citep{gasteiger2021}}} 
    & \makecell[c]{\textbf{NewtonNet} \\[-0.7ex] \scriptsize{\citep{haghighatlari2021}}}
    & \makecell[c]{\textbf{NequIP} \\[-0.7ex] \scriptsize{\citep{batzner2022nequip}}} 
    &\makecell[c]{\textbf{MD-ET (1000 samples)} \\[-0.7ex] \scriptsize{(ours, finetuned)}} 
    &\makecell[c]{\textbf{MD-ET (100 samples)} \\[-0.7ex] \scriptsize{(ours, finetuned)}}  \\
    \midrule
    Ethanol$^*$       & 15.0164 & 3.0814 & 3.0814 & 10.2424 & 2.9946 & \textbf{1.4683} $\pm$ 0.0017 & $2.5756 \pm 0.0045$ \\
    Aspirin       & 33.0274 & 10.4160 & 10.3292 & 15.4504 & 8.2894 & \textbf{4.9525} $\pm$ 0.0069 & $9.0772 \pm 1.0465$ \\
    Benzene       & 1.8228  & 0.6944  & 0.6944  & 0.4774  & \textbf{0.2604} & $2.0507 \pm 0.0645$ & $2.7354 \pm 0.0123$ \\
    Malonaldehyde & 16.2316 & 4.3400  & 5.9024  & 12.3690 & 4.5136 & \textbf{2.5632} $\pm$ 0.0158 & $5.1715 \pm 0.0688$ \\
    Toluene       & 8.8970  & 2.5172  & 2.6908  & 3.4720  & \textbf{1.6926} & $2.6117 \pm 0.0072$ & $4.3461 \pm 0.0062$ \\
    \bottomrule
\end{tabular}
}

    \label{tab:md17ccsd}
\end{table*}

MD-ET compares favorably to competing models in 3 out of 4 metrics. MD-ET's MAE score is slightly worse than that of several other models but it produces the most faithful NVT MDs according to the h(r) score (\citet{fu2022forces}). MD-ET also reaches the maximum simulation stability score across all molecules. Although a 300 ps NVT simulation is not sufficient to assess the long-term stability of MD trajectories, it can still be used to compare the stability of different models. 

\subsubsection*{MD17@CCSD(T)}
For MD simulations where precise electron correlation is necessary to accurately model potential energy surfaces and critical interactions, density functional theory (DFT) reference data may not always be sufficiently accurate. Coupled
cluster (CC) including single and double excitations with the perturbative triples correction (CCSD(T)) is often regarded as the ``gold standard'' of quantum chemistry and is thus a suitable choice for these cases. However, the substantial computational cost of CCSD(T) renders it impractical for large-scale or long-duration simulations. Transfer learning from lower-level theories, such as DFT, presents an effective strategy to approximate CCSD(T) using the expensive CCSD(T) samples efficiently \cite{kaser2023transfer}. We use the MD17@CCSD(T) dataset \cite{chmiela2018towards}, which recalculates trajectories for molecules in MD17 using CCSD(T) to test this approach using MD-ET. Although MD-ET was pretrained on the PBE0 level, Table \ref{tab:md17ccsd} demonstrates that MD-ET can effectively capture higher-level interactions through fine-tuning with a limited number of CCSD(T) samples. When fine-tuned on 1,000 CCSD(T)-level samples, MD-ET achieves lower mean absolute error (MAE) than all competing models for three of the five structures. For toluene, its MAE is comparable to GemNet-T, while for benzene, it exhibits the highest error among the benchmarked models. Furthermore, when we reduce the fine-tuning dataset by an order of magnitude MD-ET still reaches accuracy comparable to models trained on the full dataset (see Appendix \ref{sec:details-md17-ccsd} for experimental details).

\subsubsection*{Ko2020}
\label{sec:ko2020-molecule-specific-models}
\begin{table}[t]
    \caption{RMSEs of atomic forces (meV \AA$^{-1}$) for different molecular systems, comparing 4G-BPNN, SpookyNet, and MD-ET. The values for 4G-BPNN are taken from \citep{ko2021}. Best results in bold.}
    \centering
\small
\setlength{\tabcolsep}{2.5pt} %
\begin{tabular}{lcc|c}
\toprule
Molecule 
    &  \makecell[c]{\textbf{4G-BPNN} \\[-0.7ex] \scriptsize{\citep{ko2021}}}
    & \makecell[c]{\textbf{SpookyNet} \\[-0.7ex] \scriptsize{\citep{unke2021spookynet}}}
    & \makecell[c]{\textbf{MD-ET} \\[-0.7ex] \scriptsize{(ours, finetuned)}} \\
\midrule
C$_{10}$H$_2$/C$_{10}$H$_3^+$   & 78.00 &  5.802    & \textbf{5.122} $\pm$ 0.197 \\
Na$_{8}$/Cl$_8^+$               & 32.78 & \textbf{1.052}     & 2.26 $\pm$ 0.014 \\
Ag$_3^{+/-}$                    & 31.69 & 26.64     & \textbf{11.78} $\pm$ 0.409 \\
Au$_2$--MgO                     & 66.0 & \textbf{5.337} & - \\
\bottomrule
\end{tabular}

    \label{table:ko2020}
\end{table}

\citet{ko2021} introduce a molecular dataset consisting of molecular (e.g. $\text{C}_{10}\text{H}_2$ and $\text{C}_{10}\text{H}^+$) and metallic systems (e.g., $\text{Ag}_3^{+/-}$), which are typically challenging for conventional MLFFs.
Table~\ref{table:ko2020} compares the force root mean squared errors (RMSE) of MD-ET with a 4th generation Behler--Parinello neural network, 4G-BPNN (high-dimensional neural network potential, 4G-HDNNP)~\citep{ko2021} and SpookyNet~\citep{unke2021spookynet} on selected systems from the Ko2020 dataset.

For the C$_{10}$H$_2$/C$_{10}$H$_3^+$ and Ag$_3^{+/-}$ systems, MD-ET outperforms both 4G-BPNN and SpookyNet. This demonstrates MD-ET's capability to model smaller ionic and metallic systems. However, on the Na$_8$/Cl$_8^+$ MD-ET's performance is slightly worse and fails to converge on Au$_2$--MgO systems. We believe this is due to numerical instabilities in the attention layer present for larger molecules (see section \ref{sec:q2} below) as well as the distance from the training task (small molecules) to the Au$_2$--MgO system (material). 
\added{Note that our model was not adapted to periodic boundary conditions, treating the slabs instead as a finite molecule far from its training distribution.} See Appendix \ref{sec:details-ko2020} for fine-tuning details.

\begingroup
\subsubsection*{xxMD}
\begin{table*}[t]
\caption{MAE of atomic forces (meV \AA$^{-1}$) for xxMD molecules using the xxMD-DFT datasets. Best results in bold.}
\setlength{\tabcolsep}{2.5pt} %
\resizebox{\linewidth}{!}{%

\begin{tabular}{lccccccc|c}
    \toprule
    \textbf{Dataset} & 
    \makecell[c]{\textbf{MACE} \\[-0.7ex] \scriptsize{\citep{batatia2022}}} & 
    \makecell[c]{\textbf{Allegro} \\[-0.7ex] \scriptsize{\citep{musaelian2023}}} & 
    \makecell[c]{\textbf{NequIP} \\[-0.7ex] \scriptsize{\citep{batzner2022nequip}}} & 
    \makecell[c]{\textbf{SchNet} \\[-0.7ex] \scriptsize{\citep{schutt2017schnet}}} & 
    \makecell[c]{\textbf{DimeNet++} \\[-0.7ex] \scriptsize{\citep{gasteiger2020}}} & 
    \makecell[c]{\textbf{SphereNet} \\[-0.7ex] \scriptsize{\citep{liu2021}}} &
    \makecell[c]{\textbf{MD-ET} \\ [-0.7ex] \scriptsize{(ours, trained directly)}} &
    \makecell[c]{\textbf{MD-ET} \\[-0.7ex] 
    \scriptsize{(ours, finetuned)}} \\
    \midrule
    Azobenzene  & 85 & 110 & 129 & 283 & 173 & 168 & 211 & \textbf{38}   \\
    Stilbene  & 149 & 189 & 156 & 291 & 162 & 168 & 160 & \textbf{44} \\
    Malonaldehyde & 166 & 210 & 227 & 394 & 257 & 255& 257 & \textbf{36} \\
    Dithiophene  & 51 & 75 & 101 & 177 & 74 & 90 & 63 & \textbf{16} \\
    \bottomrule
\end{tabular}
}   

    \label{table:xxmd}
\end{table*}
\label{sec:xxmd}
The xxMD dataset by \citet{pengmei2024} goes beyond the MD17 dataset by incorporating nonadiabatic MD trajectories specifically designed to capture chemical reactions, including geometries sampled from the PES of reactive intermediates, transition states, and products. Importantly, xxMD datasets are also temporally split, meaning a MLFF has to generalize from a limited contiguous temporal slice to the entire trajectory.

Table~\ref{table:xxmd} shows the MAE score on xxMD. Fine-tuning for 2000 steps yields scores vastly smaller than SOTA models trained directly on xxMD. To assess how much of MD-ET's performance is due to prior knowledge obtained from the pretraining task, we also evaluate MD-ET when trained directly on xxMD (see Appendix \ref{sec:details-xxmd} for further details). Here, MD-ET performs much worse and ranks in the midfield of the other tested methods, a surprisingly strong performance given that MD-ET has no inductive biases---we hypothesize that model scale and expressiveness might positively impact generalization. Still, the difference between the directly trained and fine-tuned models demonstrates the useful physical biases attained by pretraining on a large-scale dataset like QCML.

\subsubsection*{SPICE}
All previously evaluated datasets contain conformers of a single, relatively small structure (approx. 10-20) atoms. The SPICE dataset contains several subsets with conformers of multiple structures with a size of up to 96 atoms \cite{eastman2023spice}. There are currently only two prior works reporting results on SPICE~\cite{kovacs2023mace, qu2024}, both use multiple filters to remove difficult samples from the datasets, effectively using only 88\% of the dataset \cite{kovacs2023mace}. Since our model can represent most elements up to astatine and was trained on a wide variety of structures, we use a simpler filtering mechanism and only filter examples with force vectors larger than 50~eV effectively using 98\% of SPICE (see Appendix \ref{sec:details-spice} for details). While not perfectly comparable, MAE scores of all three models are shown in Table~\ref{table:spice}. MD-ET shows competitive accuracy on all SPICE subsets,  which implies length generalization despite being trained on smaller structures -- note that, since MD-ET uses a dense attention matrix, length generalization is not a consequence of the model architecture, i.e., is a learned property.

\begin{table*}[ht]
    \centering
    \caption{MAE of atomic forces (meV \AA$^{-1}$) for the SPICE subsets. MD-ET was fine-tuned for 2000 steps. We use fewer data filtering rules, only filtering conformers containing forces over 50 eV/\AA, which results in 98\% of total samples (instead of 88\% for MACE-OFF and EScAIP, see Appendix \ref{sec:details-spice}). Results are thus only approximately comparable.}
    \setlength{\tabcolsep}{2.5pt} %
\begin{tabular}{lcc|c}
    \toprule
    \textbf{Subset} & 
    \makecell[c]{\textbf{MACE-OFF23} \\[-0.7ex] \scriptsize{\citep{kovacs2023mace}}} &
    \makecell[c]{\textbf{EScAIP} \\[-0.7ex] \scriptsize{\citep{qu2024}}} &
    \makecell[c]{\textbf{MD-ET} \\[-0.7ex] \scriptsize{(ours, finetuned, fewer filtered samples)}} \\
    \midrule
    PubChem & 14.75 & 5.86 & 11.1 \\
    DES370K Monomers & 6.58 & 3.48 & 2.55 \\
    DES370K Dimers & 6.62 & 2.18 & 4.33 \\
    Dipeptides & 10.19 & 5.21 & 4.55 \\
    Solvated Amino Acids & 19.43 & 11.52 & 8.35 \\
    \bottomrule
\end{tabular}

    \label{table:spice}
\end{table*}

\endgroup
\subsection{Unconstrained models are almost equivariant, but have systematic biases}
\label{sec:q2}
\deleted{
To further investigate MD-ET's shortcomings, including its inability to perform stable NVE simulations (see Section \ref{sec:q3}), we examine the potential influence of approximate equivariance on MD simulations more closely.
}
\added{
While MD-ET performs well on benchmarks, we want to assess the impact that the lack of architectural inductive biases has on predictions, specifically equivariance and energy conservation.
}
 To quantify how (non-)equivariant our model is, we define the equivariance error as the expected Euclidean distance between the model's mean prediction and the prediction on a randomly rotated input over all rotations in the symmetry group, 
\begin{align}
    E_\text{eq}
        ({\mathscr{D}}, \hat f_\theta)
    = \underset{\substack{x\sim{\mathscr{D}} \\ \mathcal{S}\sim\text{SO}(3)}}{\mathbb{E}} 
    \left[
        \left\|
            \underset{\mathcal{R}\sim\text{SO}(3)}{\mathbb{E}}
            \left[
                \mathcal{R}^\top  \hat f_\theta(\mathcal{R}{x})
            \right]
            - \mathcal{S}^\top
                 \hat f_\theta(\mathcal{S}{x})
        \right\|_2
    \right], 
    \label{eqn:eq_error}
\end{align}
where $x \sim {\mathscr{D}}$ are the 3D positions sampled from their distribution $\mathscr{D}$, 
    $\mathcal{R},\mathcal{S} \sim \text{SO}(3)$ are rotation matrices,
    and $\hat f_\theta$ is a trained predictor. 
For a perfectly equivariant model, $E_\text{eq}=0$ (proof in Appendix \ref{si:equiv_proof}). 
However, since models use floating point arithmetic with finite precision and numerical inaccuracies accumulate during evaluation, even equivariant models do not reach $E_\text{eq}=0$ in practice. 

When evaluating MD-ET across 2048 structures randomly sampled from the training distribution, the model incurs an equivariance error of 0.058 kcal/mol/\AA. This is several orders of magnitude lower than typical molecular forces. On the held-out test structures MD-ET's equivariance error is only slightly higher at 0.06 kcal/mol/\AA. To evaluate whether MD-ET is less equivariant on structures further from its training distribution we evaluate $E_\text{eq}$ for a range of alkanes and cumulenes of varying sizes. For comparison we also evaluate an equivariant SpookyNet with identical pretraining task and the Orb-v2 checkpoint used in the NVE experiments in the previous Section \ref{sec:md17-molecule-specific-models}. Here, all models use single-precision floating-point (fp32) arithmetic.

\begin{figure*}
    \centering
    \includegraphics[width=0.6\linewidth]{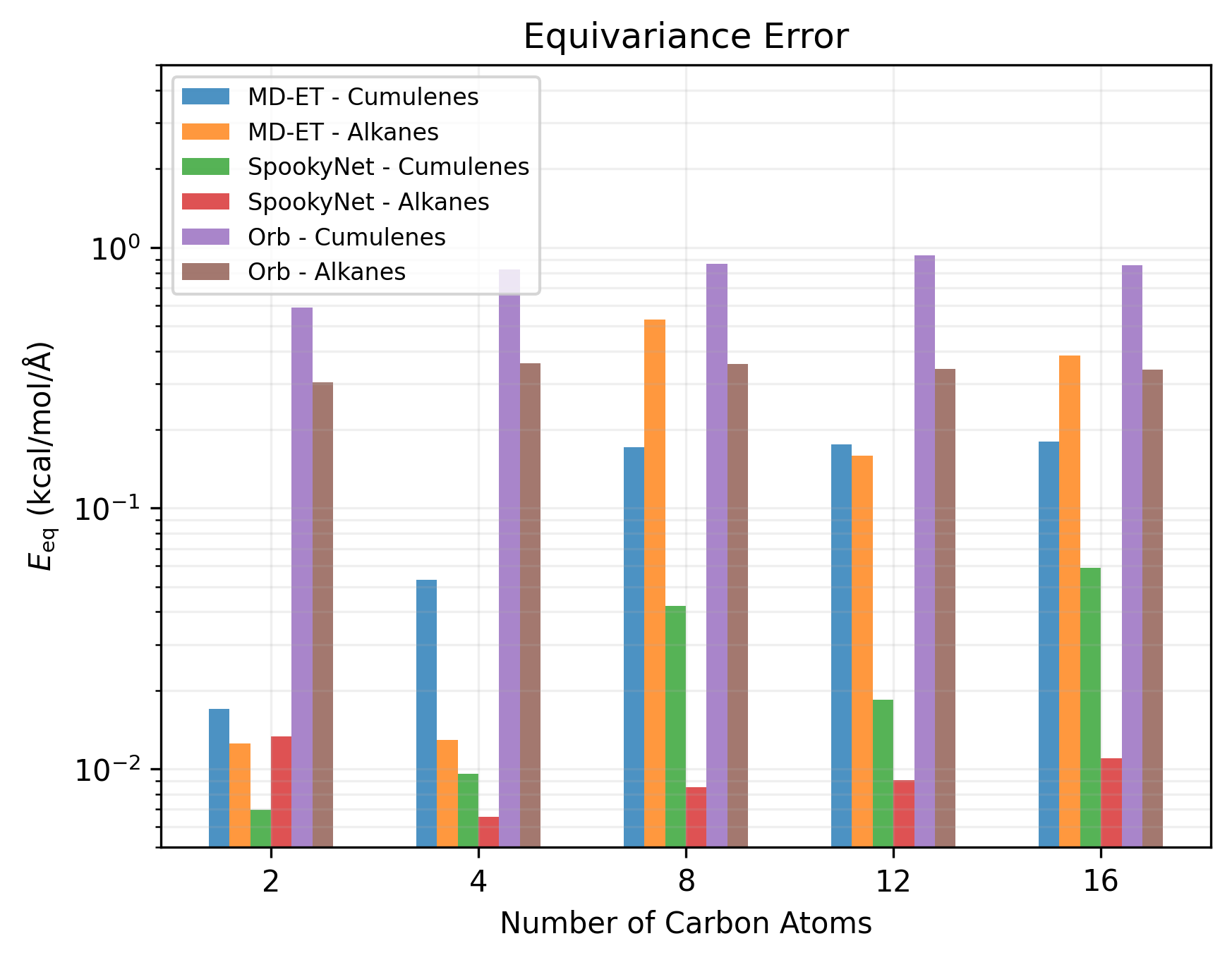}
    \caption{Equivariance error $E_\text{eq}$ over the $\mathrm{SO(3)}$ group for a range of alkanes, C$_n$H$_{2n+2}$, and cumulenes, C$_n$H$_4$ ($n=2,4,8,12,16$) measured for MD-ET (blue and orange), SpookyNet (green and red), and Orb-v2 (purple and brown), each at fp32 precision.}
    \label{fig:equivariance}
\end{figure*}

Figure \ref{fig:equivariance} shows that MD-ET's ability to predict equivariant forces diminishes outside the training distribution -- consistent with this observation we find similarly high errors for Orb across all evaluated structures. Note that Orb, which was pretrained on materials, is evaluated on molecules without further modification. It is, however notable that MD-ET's equivariance is similar to that of SpookyNet for small in-distribution structures. 
Notably, we also observe higher errors for SpookyNet on larger structures, suggesting that numerical inaccuracies, which are the only possible source of equivariance errors for SpookyNet, might also account for some of the performance degradation in unconstrained models. MD-ET's dense attention implementation might be particularly affected, as the number of attention scores which are normalized grows cubically with respect to structure size. 

Additionally, the equivariance error reports the absolute magnitude of deviations from perfect equivariance, but does not distinguish errors caused by true-to-expectation variance from those caused by systematic biases. While both kinds of errors impact $E_\text{eq}$ similarly, systematic biases are more likely to impair desirable physical properties such as energy conservation.

\begin{figure*}
    \centering
    \includegraphics[width=0.8\linewidth]{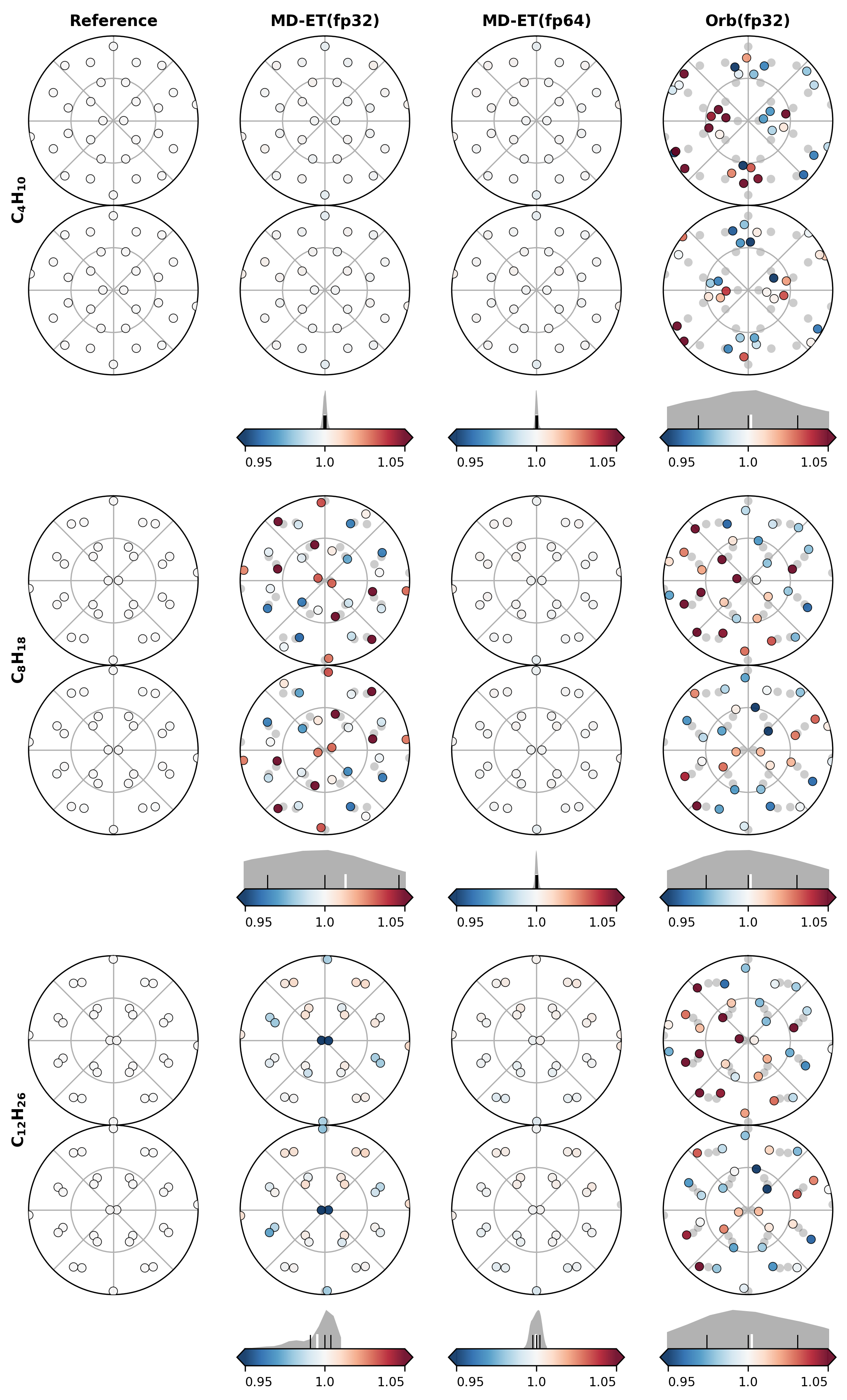}
    \caption{Qualitative analysis of equivariance errors and bias: Columns group results by method, rows group results by structure. Polar plots are 2D projections of force predictions for the second carbon atom of each structure for 60 equidistantly rotated input structures. ``Reference'' shows a perfectly equivariant PBE0 ground truth. Reference light gray ``shadows'' are shown in each plot to highlight deviations. Color coding depicts force magnitudes relative to the median prediction magnitude. Kernel density estimates visualize the distribution of force magnitudes for a larger sample of 360 equidistant rotations, where the black ticks indicate quartiles and the white tick indicates the mean.}
    \label{fig:equivariance_bias}
\end{figure*}

To further investigate the distribution of equivariance errors and the impact of numerical precision on equivariance, we visualize the force predictions for a set of rotations applied to several input structures. 
We sample 60 regularly spaced rotations via the 600-cell (Appendix \ref{app:600cell}) and use them to rotate the examined structures and visualize the resulting force predictions (see Figure \ref{fig:equivariance_bias}) using polar projections, showing the upper and lower hemispheres. 

To obtain a ground-truth reference as a point of comparison we obtain force predictions using DFT with the PBE0 functional and apply the set of rotations to it directly (column 1). These ground-truth references are additionally displayed as light gray ``shadows'' in the other columns. 
In addition to the previously examined MD-ET model (columns 2) we also evaluate the same MD-ET checkpoint using double-precision floating-point (fp64) arithmetic (column 3). Finally, we compare it again to Orb (in fp32).

In addition to angular discrepancies, which can be discerned in the projection by a mismatch of the points and the ground truth (shadows), we want to understand variability in the magnitudes of force predictions -- ideally force magnitudes should be rotation-invariant. To highlight magnitude differences we color the predictions according to their magnitude relative to the median prediction. 

Additionally, to investigate potential systematic magnitude biases, we construct a larger sample of 360 uniform rotations  (Appendix \ref{app:600cell}) and use it for a kernel density estimate (``smooth histogram'') of the force magnitude distribution below each model. 

We visualize three alkanes of different sizes. Despite being regularly spaced in SO(3) the rotation sets can lead to overlapping points when applied to vectors with certain orientations. This occurs because such an operation projects an equally spaced set of points in a 3-dimensional manifold onto a 2-dimensional manifold, i.e., force vectors that are aligned with the roll angle axis when rotating the molecule are mapped to the same position. Conversely, since rotations are regularly-spaced, points which are projected close together on the unit sphere (i.e. pitch and yaw angle are small) need to differ significantly in their roll angle, as all three need to add up to the same magnitude. If, for instance, force magnitudes of such points differ in a systematic way the model displays a roll-bias.

Although the result of applying a set of equidistant rotations to a vector are not equally distributed on the unit sphere, they are point-symmetric with respect to the origin and axis-symmetric with respect to each coordinate axis. We can therefore distinguish an error caused by a model prediction with consistent bias relative to the reference force prediction from non-equivariant predictions: A systematic bias should result in a shifted but symmetric pattern, while unbiased equivariance errors break the pattern's symmetries. Because many of the rotated reference forces of the first C atom overlap, we instead choose to visualize the forces of the second C atom for each structure.

The visualization of the predictions for these rotated inputs (see Figure \ref{fig:equivariance_bias}) shows that Orb significantly deviates from the PBE0 reference. While this is partly a result of bias, likely due to Orb being trained using data from another level of theory, we can also see that the deviations are non-symmetric, meaning Orb only imperfectly conserves orientations for the examined structures. Similarly, force magnitudes vary significantly, with Orb's having the highest variance for each sample.
For MD-ET, the results suggest that numerical stability has a significant impact on equivariance, especially for larger structures. For these structures, the min-to-max range of force magnitudes is reduced by more than a factor of ten when switching to fp64. Especially outliers seem to be mitigated by higher numerical precision, resulting in error distributions with less bias. For $\text{C}_8\text{H}_{18}$, numerical inaccuracies also lead to large angular errors, while all other examined examples are aligned so closely with the reference pattern that deviations can hardly be spotted. Overall, MD-ET's performance still appears to deteriorate as structures become larger. Notably, even as higher numerical precision removes random orientation-based fluctuations in force magnitudes, MD-ET displays a small but clear directional bias, one half-space of orientations resulting in significantly higher force predictions. Although these deviations are minor and unlikely to impact observables in an NVT simulation, they are likely to lead to instability in NVE simulations, because the effect of minor errors accumulates over time. Orb displays a similar magnitude bias for the top left quadrant. Furthermore, we can see that force magnitudes predicted by Orb also significantly differ for pairs of input rotations with similar pitch and yaw angles, but different roll angle.

\subsection{Unconstrained can models learn to approximately conserve energy without explicit inductive biases}
\label{sec:q3}
\begin{figure}[ht]
\centering
\includegraphics[width=0.8\linewidth]{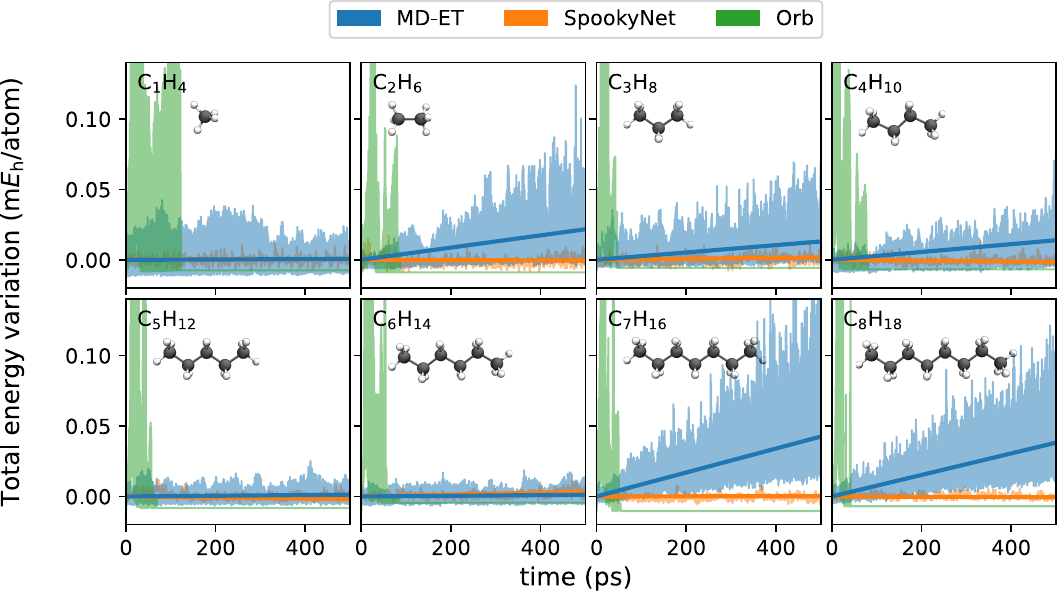}
\caption{Total energy variation per atom over time for NVE simulations of linear alkanes C$_n$H$_{2n+2}$ ($n=1\dots8$). For reference, we compare our model (blue) to MD simulations with an energy-conserving model, SpookyNet \cite{unke2021spookynet} (orange) trained on the same data as MD-ET, and the public checkpoint of another non-conserving model, Orb-v2 \cite{neumann_orb_2024} (green), trained on periodic systems. A linear fit is plotted for the first two to help visualize energy drift.}
\label{fig:energy_conservation}
\end{figure}
Since MD-ET directly predicts forces without enforcing energy conservation, we aim to assess whether the predicted forces are approximately conservative. Although MD-ET achieves stable NVT MD simulations for MD17 molecules (see Section \ref{sec:md17-molecule-specific-models}), this alone is not sufficient to demonstrate approximate energy conservation. This is because the dynamics sample the canonical (NVT) ensemble, i.e., the aim is to keep the temperature of the simulated system constant. To achieve this, a thermostat constantly introduces or removes energy, which prevents testing energy conservation. We therefore also perform MD simulations sampling the microcanonical (NVE) ensemble, where total energy is a conserved quantity. This enables quantification of how strongly a simulation violates energy conservation by monitoring the variation of total energy over time. Note that even when using conservative MLFFs, the total energy typically fluctuates slightly due to numerical noise and discretization errors when integrating the equations of motion. This is acceptable as long as these fluctuations are small and stay centered around zero, i.e., do not introduce energy drift. 

We find that NVE simulations with MD-ET are approximately energy-conserving for some small systems, but with increasing system size the total energy tends to increase with time (see Figure \ref{fig:energy_conservation}). The additional energy causes increasingly large structural fluctuations until the simulation invariably becomes unstable after a sufficiently large number of time steps. To compare the MD-ET results and assess how out-of-distribution inference of unconstrained models affect energy conservation, we also assess NVE stability using the publicly available checkpoint of Orb-v2 \cite{neumann_orb_2024}. Although Orb  has been pretrained on periodic structures, it has shown promising zero-shot results on MD17-10k molecules with $h(r)$ scores similar to the ones reported for MD-ET above (see Table \ref{tab:hyp_md17}). However, in our NVE simulations Orb seems to approximate energy conservation much less closely than MD-ET, failing to keep the simulation stable for even short durations. These findings agree with \citep{bigi2024}, who found that Orb is less energy conserving than other unconstrained models. \deleted{To find out what causes the big discrepancy between Orb's performance and that of MD-ET we investigate their approximate equivariance next.}

\begin{figure}[ht]
\centering
\includegraphics[width=0.8\linewidth]{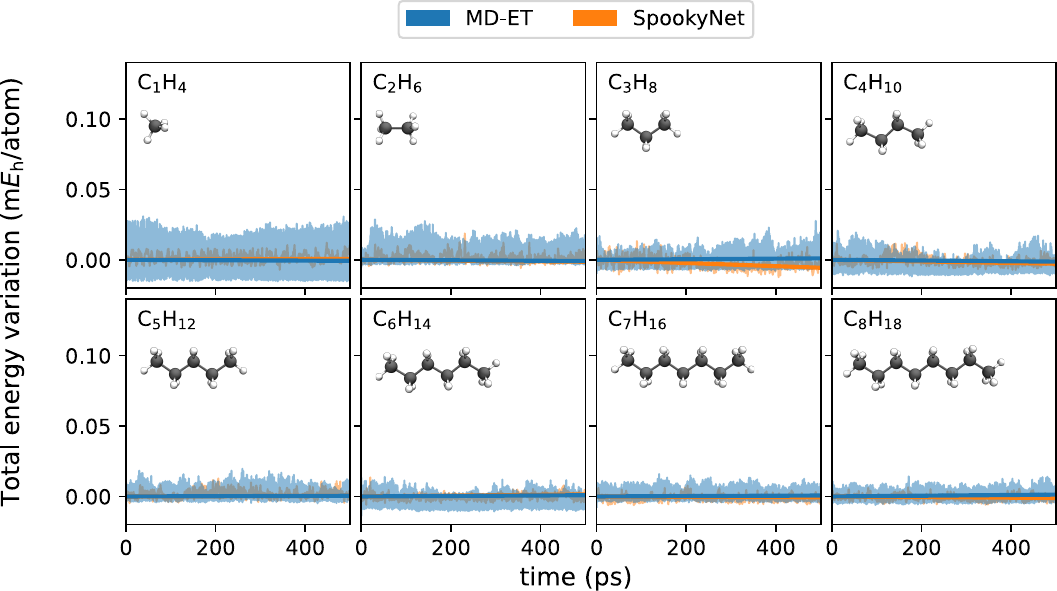}
\caption{Total energy variation per atom over time for NVE simulations of linear alkanes C$_n$H$_{2n+2}$ ($n=1\dots8$), same as depicted in Figure \ref{fig:energy_conservation}. However, both models were evaluated using fp64 precision instead of fp32  precision, MD-ET additionally samples a random offset rotation for the input structure at each step.}
\label{fig:energy_conservation_fp64}
\end{figure}

\begin{figure}
    \centering
    \includegraphics[width=0.8\linewidth]{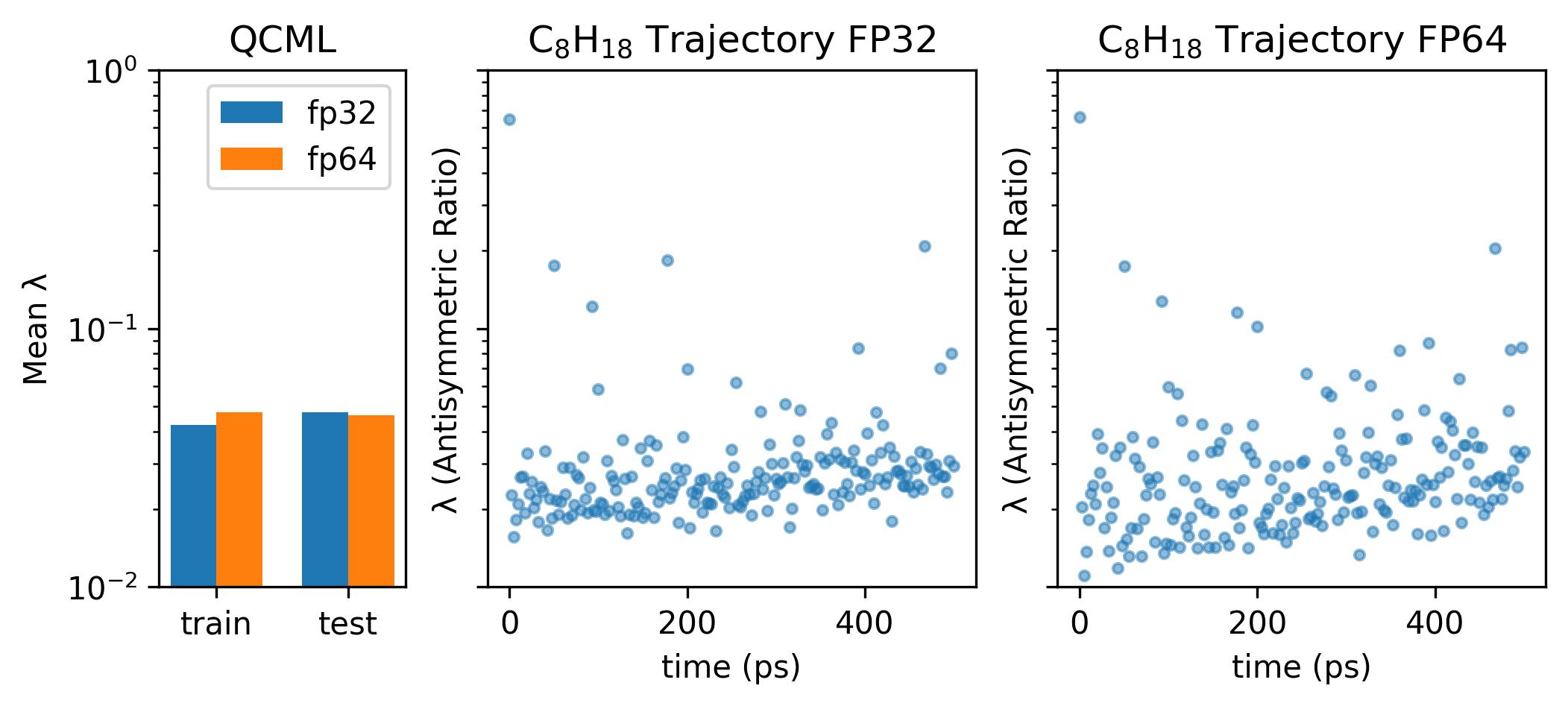}
    \caption{Left: Mean Antisymmetric Ratio $\lambda$ for 2048 samples of QCML train and test set, evaluated using fp32 and fp64 precision. Middle/Right: Antisymmetric ratios $\lambda$ evaluated on frames from an unstable fp32 NVE trajectory - $\lambda$ values evaluated using fp32 (middle) and fp64 (right) precision.}
    \label{fig:energy_conservation_lambda}
\end{figure}

\begin{figure}
    \centering
    \includegraphics[width=0.7\linewidth]{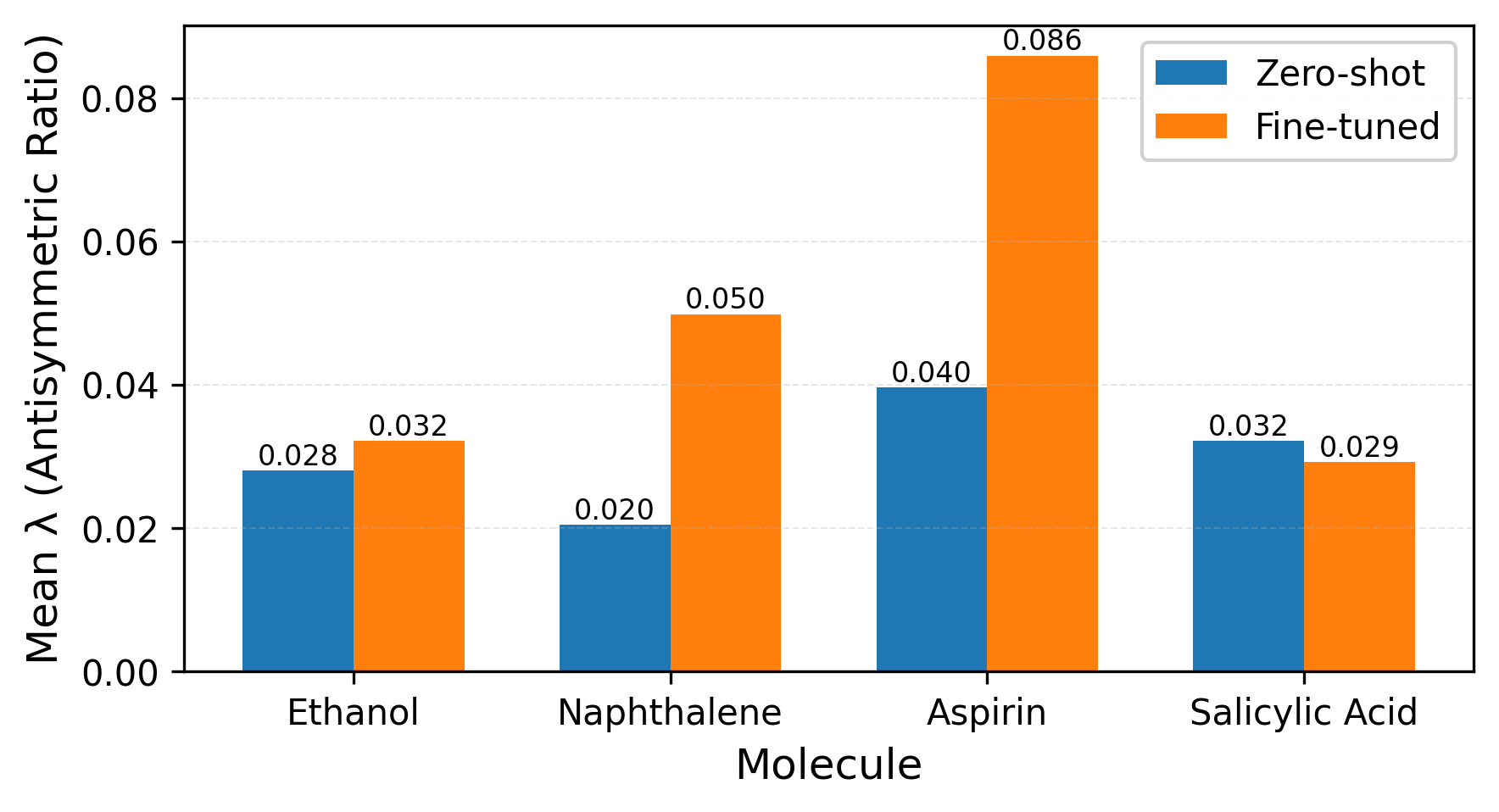}
    \caption{Effect of fine-tuning on Antisymmetric Ratio $\lambda$, mean of $\lambda$ for 256 frames of MD17 before and after fine-tuning. Fine-tuning and evaluation using fp64 precision to eliminate the effect of numerical imprecision.}
    \label{fig:energy_conservation_ft_md17}
\end{figure}

\deleted{To validate the findings of the previous section we run additional NVE simulations.} 
\added{However, similarly to energy conservation, we find that we can stabilize NVE simulations significantly when evaluating the model using fp64 precision.} Of the eight alkanes tested in Section \ref{sec:q2}, Figure \ref{fig:energy_conservation}, six remain stable when evaluating the model using fp64. To test whether the minimal remaining bias visible in Figure \ref{fig:equivariance_bias} has an impact on the simulations, we add a random offset rotation to the simulation, i.e., at each simulation step we sample a random 3D rotation and apply it to the positions. \added{This approach was also discussed by \cite{langer2024} as a method to improve equivariance.} We then apply the reverse rotation to the force predictions. This procedure has the benefit of removing the autocorrelation of rotational biases from the simulation, allowing stable NVE simulations for all previously considered alkanes (see Figure \ref{fig:energy_conservation_fp64}). 

\added{To clarify these results we introduce a second metric for assessing the energy conservation of model predictions. \citet{bigi2024} examine the Jacobian matrix of the force prediction with respect to atom positions $\mathbf{J} \in \mathbb{R}^{3n \times 3n}$ to quantify energy conservation. The Jacobian of a perfectly energy-conserving prediction should be a symmetric matrix since it is the Hessian of the molecular energy. \citet{bigi2024} quantify the symmetry of $\mathbf{J}$ by decomposing the matrix into its symmetric and antisymmetric components and calculating the ratio between the Frobenius norm of the antisymmetric component and the Frobenius norm of the entire matrix, i.e,
\begin{equation}
    \lambda = \frac{\|\mathbf{J}_{\text{anti}}\|_F}{\|\mathbf{J}\|_F},
\end{equation}
where $\mathbf{J}_{\text{anti}} = (\mathbf{J} - \mathbf{J}^T)/2$. Thus $\lambda$ takes a value between zero and one, where zero represents a perfectly symmetric Jacobian and a matrix with independent values randomly sampled from $\mathcal{N}(0,1)$ has an expected value of $\lambda = \frac{1}{\sqrt{2}} \approx 0.7 $ for large $n$.
}

\added{
To understand how energy-conserving MD-ET is and how using fp64 precision impacts energy conservation, we first evaluate $\lambda$ for 2048 conformer samples of QCML's training and test sets using fp32 and fp64 precision (Figure \ref{fig:energy_conservation_lambda}, left). We also collect frames from the unstable fp32 MD trajectory of C$_8$H$_{18}$ and evaluate $\lambda$ on them using both fp32 and fp64 precision (Figure \ref{fig:energy_conservation_lambda}, middle/right).
}

\added{
Results show that MD-ET is not only able to run stable NVE simulations for specific structures but generally learns to make predictions that are approximately energy conserving. While evaluating using fp64 precision is slightly detrimental on training samples and has no significant impact on the QCML test set, it seems to impact predictions on actual NVE trajectories. The overall impact there is nevertheless relatively small, i.e., fp64 precision is detrimental for some frames but reduces $\lambda$ for some by 20--35\%.
} 

\added{Stable NVE simulations appear to be harder for larger structures. However, $\lambda$ is slightly lower for larger structures of QCML (see Figure \ref{fig:lambda_vs_size_qcml}, Appendix \ref{sec:sup_figures}). We therefore believe that simulations of larger structures tend to be more sensitive to accumulating errors.}

\added{
Since we have demonstrated earlier that MD-ET can be fine-tuned efficiently, we also investigate the effect of fine-tuning on energy conservation. We use the MD17 fine-tuned checkpoints of MD-ET from the benchmark section as well as the QCML-pretrained model and evaluate both on 256 test set frames of each molecule (see Figure \ref{fig:energy_conservation_ft_md17}). We observe that fine-tuning slightly degrades energy conservation, suggesting that large-scale pretraining on QCML is more beneficial for learning energy conservation than training on smaller datasets. 
}

\added{
In summary, these results suggest that the long-term stability of NVE simulations is sensitive even to small deviations from perfect energy conservation (which do not cancel over time). In this section we have shown two methods to further reduce such systematic errors. Evaluating MD-ET using fp64 precision appears to reduce errors on relatively few out-of-distribution frames, which have a significant impact on NVE stability. Furthermore, adding random rotational offsets eliminates the autocorrelation of directional biases incurred through equivariance errors. We assume such biases can reinforce or dampen oscillations that traverse several different orientations, accumulating or dispersing energy. Using these two methods, MD-ET is able to produce stable NVE simulations for a number of structures, even outside the training manifold (see Figure \ref{fig:nve_additional_structs}, Appendix \ref{sec:sup_figures}). MD-ET, however, fails for many other structures. While values for $\lambda$ differ between structures, we also believe that structures are sensitive to error accumulation to different degrees; for instance, in our testing only very small cumulenes tend to remain stable.
}

\added{We have also shown in line with \citet{bigi2024} that rates of energy `leakage' and thus heating during NVE simulations varies significantly between models.}

\deleted{
We draw two conclusions from this: 
First, MD-ET has learned approximate equivariance for some of the tested molecules without an explicit inductive bias or further fine-tuning on tested structures. 
While all eight alkanes shown here are part of the QCML dataset, $\text{C}_9\text{H}_{20}$ and $\text{C}_{10}\text{H}_{22}$ are not; nevertheless, MD-ET is capable of running stable NVE simulations on them even without random offsets (see Figure \ref{fig:nve_additional_structs} in Appendix \ref{sec:sup_figures}).
However, MD-ET is still unable to run stable NVE simulations for many other structures, such as several cumulenes, which have complex long-range interactions. 
Second, since NVE accumulates errors over many steps, even small deviations such as small numerical instabilities (which introduce bias) or prediction biases for some orientations make an unconstrained model ``explode'', even if it is approximately energy-conserving. We think it is thus very likely, that, in practice, such a model needs to be very close to equivariant to be able to conserve energy. Although fp64 precision can remove numerical instabilities and random offset rotations effectively prevent rotational biases from impacting energy conservation, both have downsides: The former introduces computational overhead (which might be greater than that of predicting conservative forces) while random offset rotations introduce a small amount of noise with a variance equivalent to $E_\text{eq}$ to the simulation.
}

\deleted{
While numerical stability and equivariance are important for energy conservation, the most significant factors for energy conservation are still model architecture and training protocol. We highlight this in Appendix Figure \ref{fig:nve_traj_comparison} (see Appendix \ref{sec:sup_figures}): Even at comparable levels of equivariance error, the potential energy of the system simulated by Orb grows more than 10 times faster than for MD-ET. 
}

\subsection{Unconstrained models are able to reproduce observables in NVT simulations}
\begin{figure}
    \centering
    \includegraphics[width=0.8\linewidth]{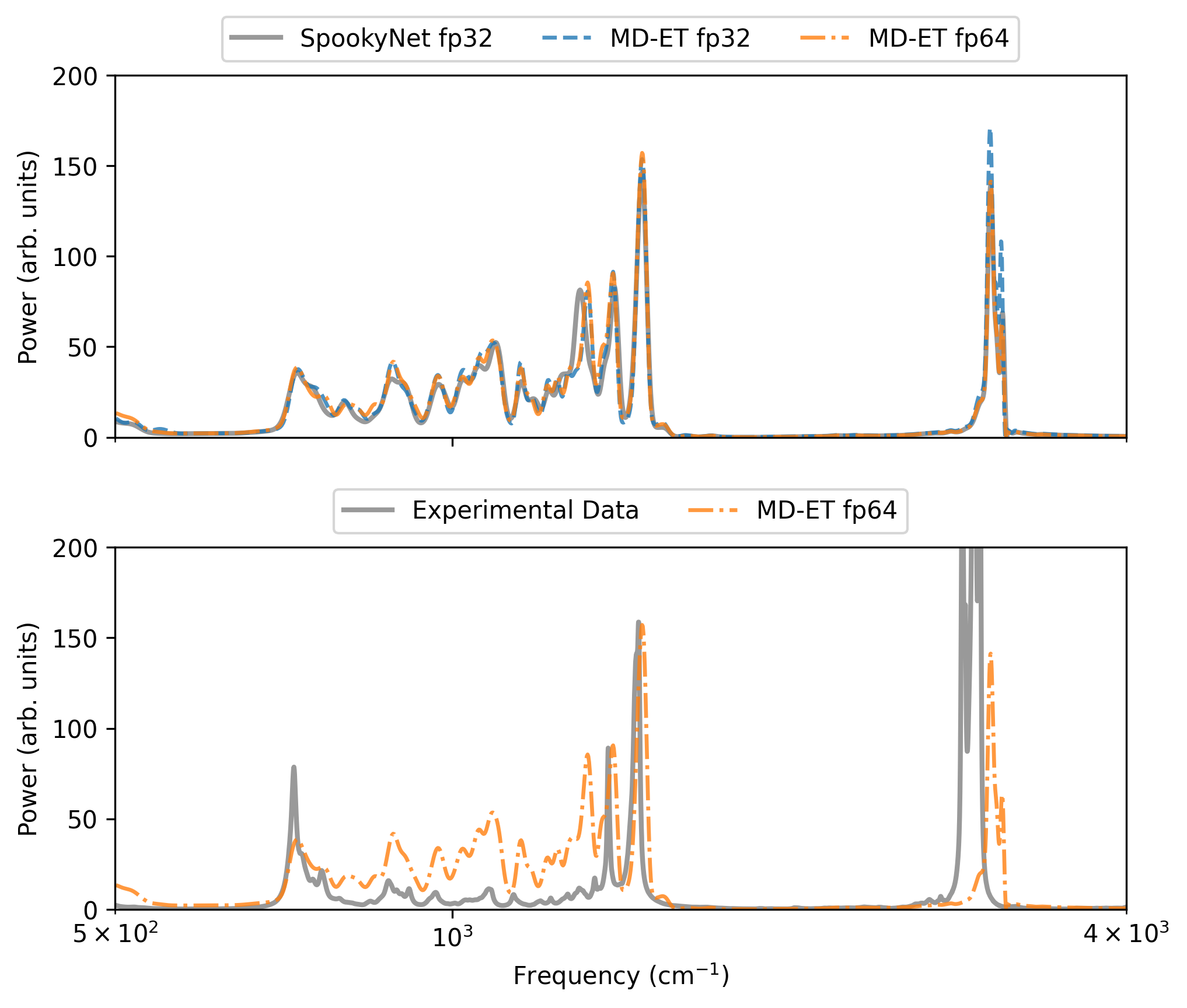}
    \caption{Top: Velocity-velocity autocorrelation power spectra of a 1~ns simulation of $n$-octane, $\text{C}_8\text{H}_{18}$, using the SVR thermostat with $\tau=1000$. MD-ET and SpookyNet are identically trained and MD-ET is evaluated using fp32 and fp64. Bottom: Comparison of MD-ET fp64 spectrum to an experimentally observed infrared absorption spectrum \cite{Myers2025IARPA,NISTWebBook2025}.}
    \label{fig:faithfulness_1}
\end{figure}
\deleted{While results discussed in Section \ref{sec:q1} show that unconstrained models are competitive in benchmark accuracy, speed and NVT simulation stability, Sections \ref{sec:q2} and \ref{sec:q3} examine significant problems with conserving forces, leading to unstable NVE simulations. The unphysical predictions leading to this behavior are undesirable. However, it is an open question how harmful these, at times very small, deviations from physics are to the overall simulation faithfulness of such models and whether they can be mitigated with a thermostat.}
\added{
 In the previous section we have shown that MD-ET tends to make predictions that are approximately energy conserving. However, even small systematic errors can catastrophically impact NVE simulations over time. Unfaithfulness to energy conservation can be quantified as a ``drift rate'', which differs between models and structures. This section tries to answer the question of whether, given a desired observable, valid results can be derived despite a non-zero energy drift rate.
}

The $h(r)$ scores reported in Table \ref{table:md17} suggest that simulation quality is not affected provided a suitable thermostat is used, with both Orb and MD-ET reaching competitive scores. However, the $h(r)$ measure is not directly motivated by physics and thus difficult to interpret. Furthermore, as \cite{bigi2024} discusses, the choice of thermostat is important when comparing observables. 

To evaluate the degree to which MD-ET produces faithful MD simulations we run 1~ns NVT simulations using the stochastic velocity rescaling (SVR or Bussi) thermostat \cite{Bussi2007}, which rescales all atom momenta by a stochastically sampled scaling variable. Importantly, when the temperature slightly deviates from the target temperature and the friction coefficient $1/\tau$ is low, the Bussi thermostat reproduces the desired ensemble, conserving important quantities. We choose $n$-octane, $\text{C}_8\text{H}_{18}$, as the structure for the experiment, where the fp32-evaluated MD-ET shows significant equivariance and energy conservation errors, while the fp64-evaluated MD-ET is energy conserving in the NVE ensemble (see Figure \ref{fig:nve_traj_comparison} in Appendix \ref{sec:sup_figures}). We choose a low friction coefficient with $\tau = 1000$ to make sure to capture any artifacts produced by equivariance and energy conservation errors and obtain trajectories from MD-ET (both fp32- and fp64-evaluated) and an equivalently trained SpookyNet. We then produce velocity-velocity autocorrelation spectra from these trajectories (see Figure \ref{fig:faithfulness_1}, top). Note that both SpookyNet and the fp64-evaluated MD-ET are energy-conserving and (very close to) equivariant for $\text{C}_8\text{H}_{18}$. While all three trajectories are nearly identical, we can see minor differences between all of them. The only notable differences are the more pronounced high-frequency vibrations only produced by fp32-evaluated MD-ET and a slightly shifted mid-frequency peak relative to SpookyNet caused by both MD-ET models. 
In Figure \ref{fig:faithfulness_1}, bottom, we compare the same fp64-evaluated MD-ET spectrum with the gas-phase IR spectrum of $n$-octane taken from the NIST/EPA vapor-phase library \cite{Myers2025IARPA, NISTWebBook2025}. The slight red-shift of the experimental spectrum relative to MD-ET is likely due to temperature, quantum--classical differences, instrumental calibration, and the residual error of MD-ET.

\added{
To more directly compare the effect of energy conservation on observables we train an identical ET model utilizing energy-conserving forces through automatic differentiation and compare a 50\,ns (zero-shot) free energy surface of alanine dipeptide for both this energy-conserving model and MD-ET. We also examine the effect of using random rotational offsets for each MD frame. The results indicate no significant difference between all three free energy surfaces (see Appendix \ref{app:alanine-fes}).
}

In the above NVT results as well as the free energy surfaces we do not see equivariance and energy conservation errors introduced by MD-ET affect observables. However, these results might not extend to models such as Orb (see results for Orb in Appendix \ref{sec:sup_figures}, Figure \ref{fig:faithfulness_1_orb}), which has more severe equivariance and energy conservation errors as well as to more complex structures, $\text{C}_8\text{H}_{18}$ being in the training dataset \added{(alanine dipeptide is not)}. Although we cannot draw general conclusions from such a limited evaluation, we think it is reasonable to assume that NVT results of unconstrained models can be relied upon given their ``energy drift rate'' is below some threshold. A theoretical argument for this hypothesis is that the SVR thermostat samples the canonical ensemble when in equilibrium, i.e., when the energy drift rate is close to zero and the temperature thus stable. \added{Since some thermostats such as SVR allow to track such quantities, unconstrained models might be used to predict some observables in practice, given this quantity is monitored and within some acceptable range.} Nevertheless, a more principled and systematic evaluation which is beyond the scope of this work is needed to give conclusive evidence.

\section{Conclusion}
\label{sec:conclusion}   

There is an ongoing debate in the MD community whether including inductive biases such as energy conservation or rotational equivariance in ML architectures is necessary for successful MD applications. ML methods have always needed to find a balance between ease of optimization and the need to include helpful inductive biases. While traditional models for MD lean toward strong inductive biases, here, we present MD-ET, an edge transformer-based approach to MD simulations that relaxes strict equivariance and energy conservation constraints during training. While not meant as a serious general-purpose platform due to its dense attention and cubic scaling, we use MD-ET as a simple and expressive test platform to evaluate the promise of fully unconstrained models for MD.

For a minimally MD-adapted architecture, MD-ET shows surprisingly good benchmark results, as well as good few-shot transfer and zero-shot capabilities, including stable few-shot MD simulations in the NVT ensemble. To better understand potential problems with unconstrained architectures, we experimented with MD-ET, studying potentially problematic properties. \deleted{With regards to approximate energy conservation, we find that MD-ET is capable of learning to conserve energy and run stable NVE simulations for some structures in and close to the data distribution without explicit inductive biases or training protocols. However, this property differs from structure to structure and is sensitive to small biases such as equivariance or numerical noise. Evaluating equivariance, we found that MD-ET has learned to predict approximately equivariant forces with minimal training interventions. Although performance here is much more consistent than for energy conservation, small variations between structures and small directional biases remain. Regarding the use of unconstrained models to derive observables in the NVT ensemble, we find hardly any notable differences between MD-ET and SpookyNet in our limited testing, even on structures which exhibit instabilities in the NVE ensemble.
}
\added{
 Evaluating equivariance, we found that MD-ET has learned to predict approximately equivariant forces with minimal training interventions. However, small variations between structures and small directional biases remain. With regards to approximate energy conservation, we find that MD-ET learns to make approximately energy-conserving predictions and can run stable NVE simulations for some structures in and close to the data distribution without explicit inductive biases or training protocols. However, this property differs from structure to structure and is sensitive to small biases such as equivariance or numerical noise. Regarding the use of unconstrained models to derive observables in the NVT ensemble, we find hardly any notable differences between models with and without strict inductive biases in our limited testing, even on structures which exhibit instabilities in the NVE ensemble.
}

MD-ET's performance suggests that models for MD simulations may not necessarily need a special-purpose architecture to adhere to physical constraints, and instead, these constraints can also be learned from data. This is particularly successful for rotational equivariance, which MD-ET can learn up to several orders of magnitude below typical force magnitudes. \deleted{Whether energy conservation can similarly be learned consistently and whether non-conservative MLFFs can be used for reliable MD simulations of large molecular systems is questionable, at least without further advances, such as additional loss terms that encourage learning of energy-conserving forces. }
\added{
Whether energy conservation can similarly be learned consistently enough to allow reliable MD simulations of large molecular systems is questionable, at least without further advances, such as additional loss terms that encourage learning of energy-conserving forces. 
}

\deleted{
The main limitation of our study is the lack of quantitative evaluation metrics to test for failure modes of unconstrained MLFFs. Our results demonstrate a disagreement between commonly used quantitative benchmark metrics and more qualitative but targeted evaluation. While we were able to identify some of the failure modes of unconstrained models, a larger set of targeted, quantitative tests is needed to conclusively answer all of the posed questions. Since unconstrained architectures do not guarantee properties such as equivariance or energy conservation for all inputs, such comprehensive evaluation is especially important for these methods. 
}

 In summary, we studied possible limitations that may result from unchaining all architectural and physical constraints from MD models. Our proposed MD-ET is surprisingly competitive on benchmarks, approximately equivariant, and can perform some stable NVE simulations. However, our results also suggest that it is difficult for unconstrained models to learn approximate energy conservation \added{to the degree necessary to consistently achieve stable NVE simulations}, and that MD simulations driven by non-conservative forces should be carefully assessed on a case-by-case basis. We consider the large corpus used for pre-training our model to be a key ingredient for MD-ET's accuracy, along with the comparative ease of optimization of Edge Transformers. We would like to emphasize the limits of any empirical study (including ours) and will in the future strive for further theoretical insights and a more quantitative empirical evaluation of the failure modes of unconstrained models.

\section*{Software and data}
Project code is available on GitHub: \href{https://github.com/mx-e/simple-md}{https://github.com/mx-e/simple-md}. For data, see \citet{ganscha2025qcml}.

\section*{Supplementary Material}
\added{
See the supplementary material for additional details on the model architecture and implementation, comprehensive training protocols and hyperparameters, dataset splits and preprocessing, and extended results including further analysis of equivariance errors, energy conservation metrics, and computational performance.
}

\section*{Acknowledgements}
We thank Luis M\"uller for his encouragement and being a great colleague. 
We also thank Elron Pens, Stefan Chmiela, and Sidney Bender for important insights and fruitful discussions.
SGu was supported by the Postdoc.Mobility fellowship of the Swiss National Science Foundation (project no.\ 225476).
ME, SGu, and KRM acknowledge support by the German Ministry of Education and Research (BMBF) for BIFOLD (01IS18037A). Further, this work was in part supported by the BMBF under Grants 01IS14013A-E, 01GQ1115, 01GQ0850, 01IS18025A, and 031L0207D. KRM was partly supported by the Institute of Information \& Communications Technology Planning \& Evaluation (IITP) grants funded by the Korea government (MSIT) (No.2019- 0-00079, Artificial Intelligence Graduate School Program, Korea University and No. 2022-0-00984, Development of Artificial Intelligence Technology for Personalized Plug-and-Play Explanation and Verification of Explanation).

\putbib[bib]
\defaultbibliographystyle{unsrtnat}
\end{bibunit}

\newpage
\appendix
\onecolumn
\begin{bibunit}
\section{Appendix}
\setcounter{page}{1}

\subsection{Supplementary Figures}
\label{sec:sup_figures}
\begin{figure}[ht!]
    \centering
    \includegraphics[width=0.8\linewidth]{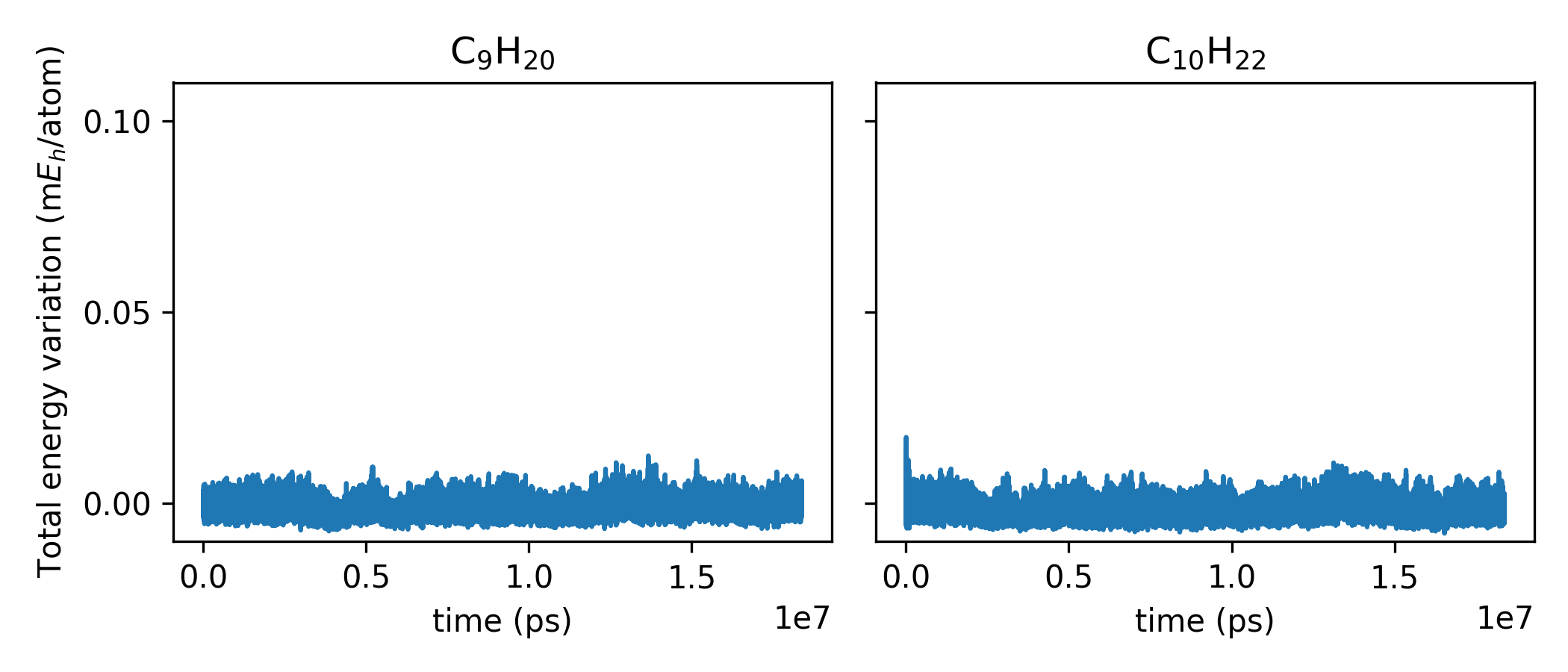}
    \caption{Total energy variation per atom over time for NVE simulations of additional alkanes, using MD-ET evaluated with fp64 precision.}
    \label{fig:nve_additional_structs}
\end{figure}

\begin{figure}[ht!]
    \centering
    \includegraphics[width=0.85\linewidth]{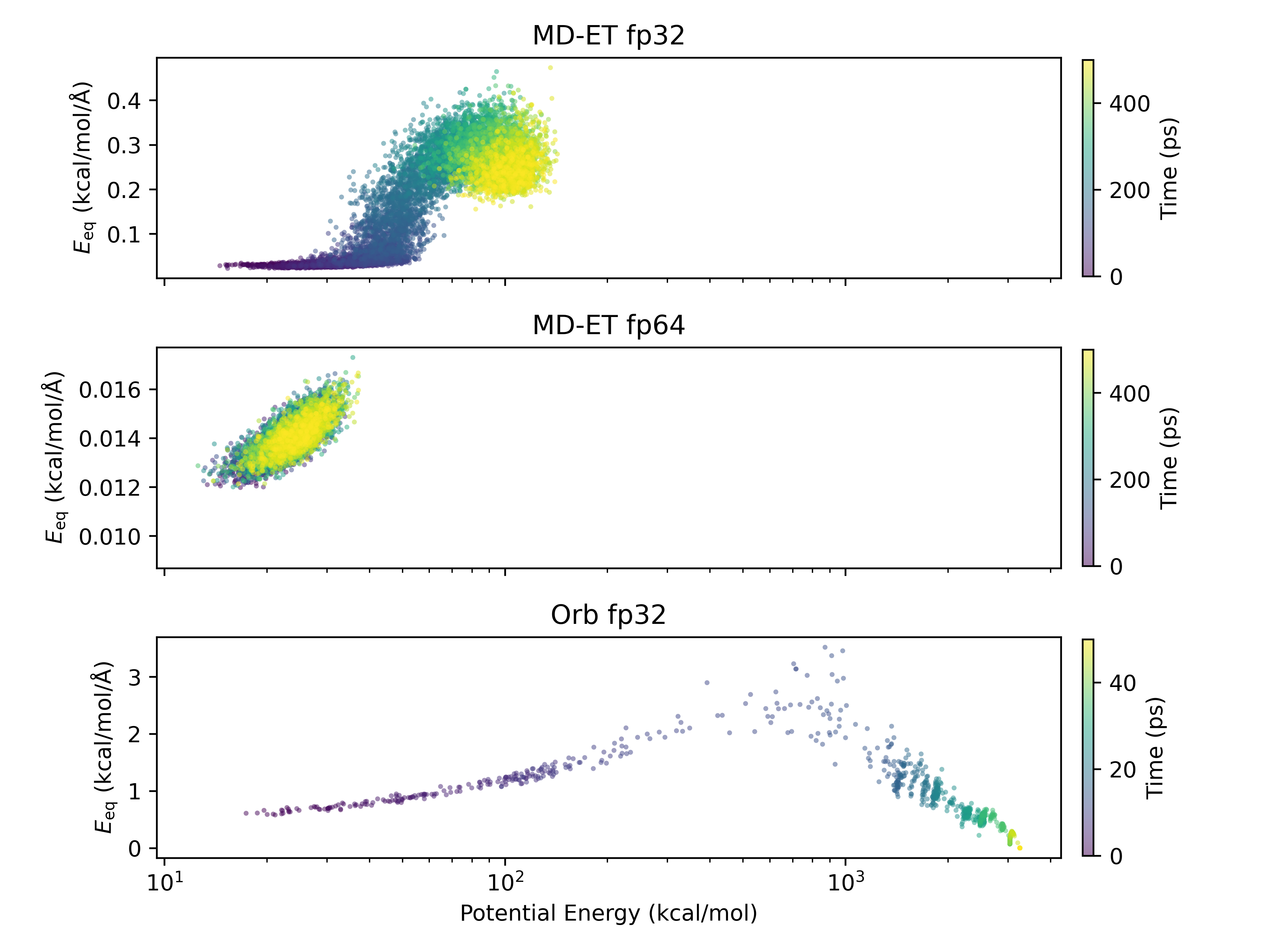}
    \caption{Development of equivariance error $E_\text{eq}$ and potential energy along a 500 ps NVE trajectory simulating $\text{C}_8\text{H}_{18}$ sampled every 50~fs. Comparison between MD-ET (identical checkpoint, fp32 and fp64 evaluated) and Orb-v2. The Orb plot only shows the first 50 ps since the molecule ``explodes'' after, which makes energy calculations unreliable. Potential energy of frames calculated with PBE def2-SVP.}
    \label{fig:nve_traj_comparison}
\end{figure}
\begin{figure}[ht!]
    \centering
    \includegraphics[width=0.7\linewidth]{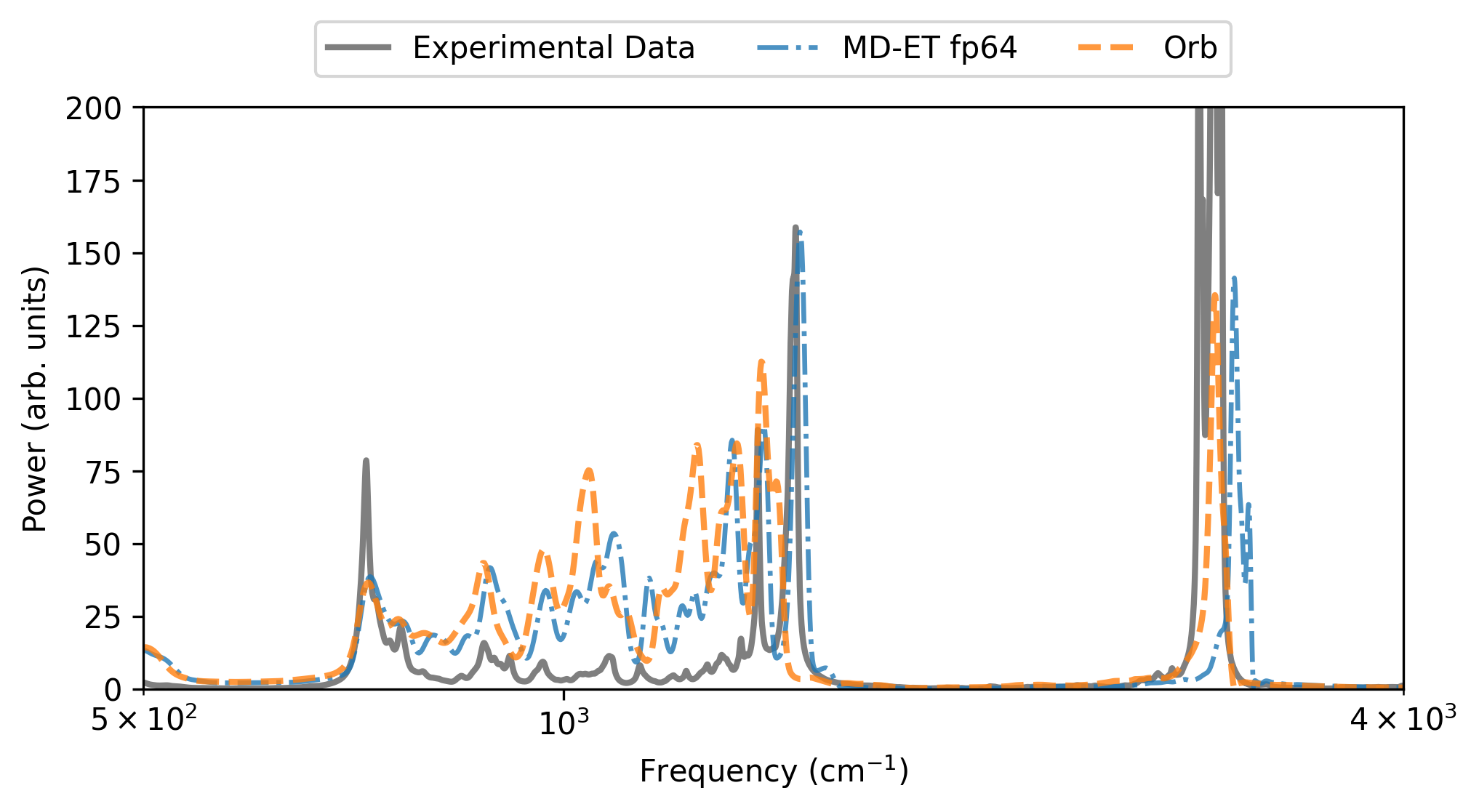}
    \caption{Velocity-velocity autocorrelation power spectra of a 1~ns simulation of $\text{C}_8\text{H}_{18}$ using the SVR thermostat with $\tau=1000$. Comparison of Orb and MD-ET (fp64 evaluated) with an experimentally observed infrared absorption spectrum \cite{Myers2025IARPA,NISTWebBook2025}}
    \label{fig:faithfulness_1_orb}
\end{figure}

\begin{figure}
    \centering
    \includegraphics[width=0.7\linewidth]{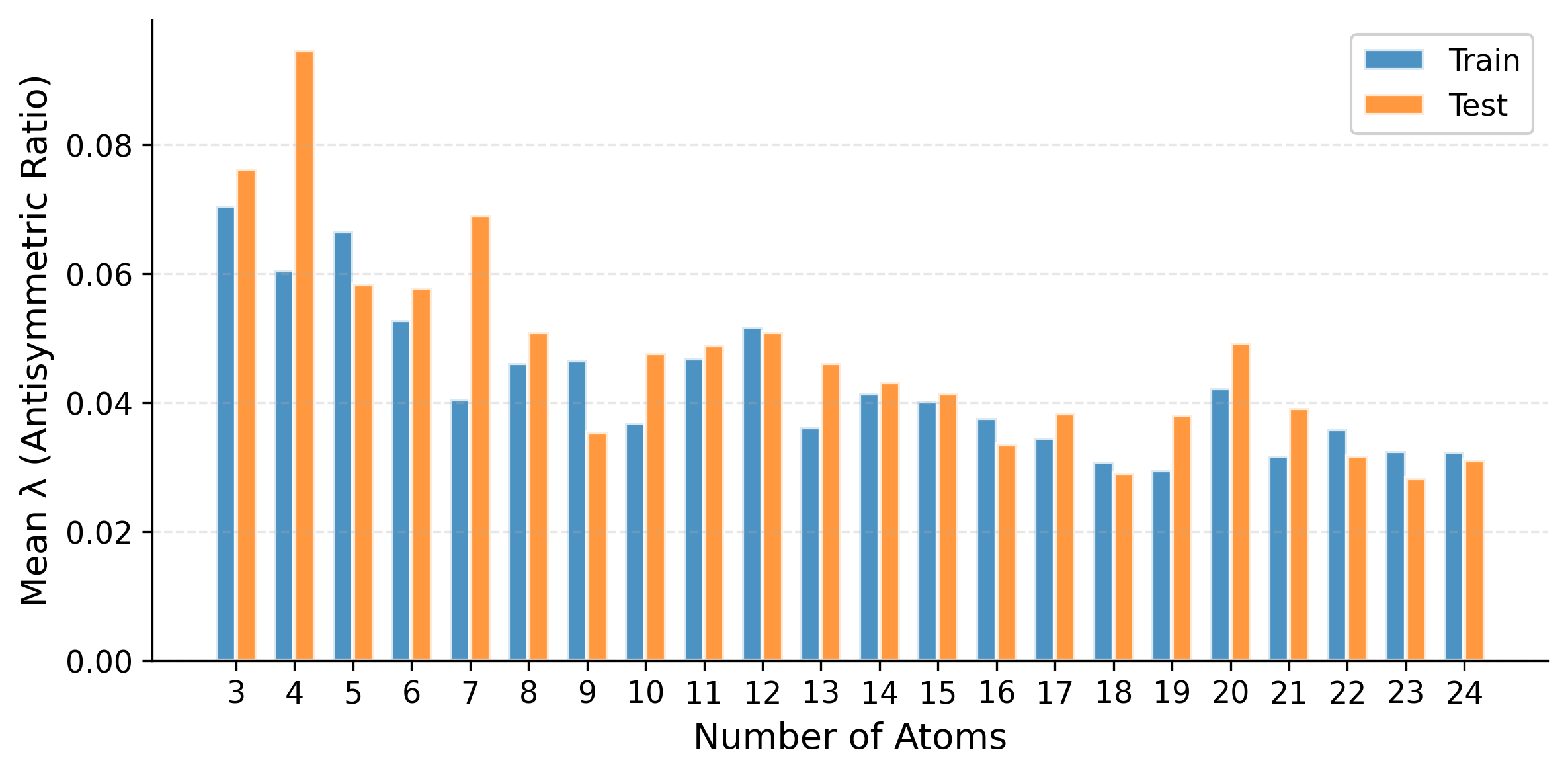}
    \caption{\added{Mean antisymmetric ratio $\lambda$ of force predictions of 2048 QCML train/test samples. Sizes smaller than 2 atoms or with less than 10 samples have been omitted.}}
    \label{fig:lambda_vs_size_qcml}
\end{figure}
\newpage
\subsection{Molecular embedding layer implementation details} 
\label{sec:embedding_layer_details}

To create initial edge representations, we combine several embeddings: i) spin and charge, ii) atomic numbers, iii) pairwise distances, and iv) pairwise directions. 

\textbf{Spin and charge} are embedded through learned embedding layers. The total spin, $s_i \in \mathbb N_0$, denotes the number of unpaired electrons, and $q_i \in \mathbb Z$ denotes the net charge. They are jointly embedded as
\begin{equation}
    \mathbf{e}_{i} =
        \mathbf{W}^\text{spin} {h}(s_i)
        + \mathbf{W}^\text{charge} {h}(q_i),
\end{equation}
where
$\mathbf{W}^\text{spin} \in \mathbb{R}^{D \times d_\text{spin}}$ 
and
$\mathbf{W}^\text{charge} \in \mathbb{R}^{D \times d_\text{charge}}$
are learned weight matrices for spin and charge, respectively, ${h}$ is a one-hot encoding and $d_\text{spin}$ and $d_\text{charge}$ are the sizes of the spin and charge vocabularies, i.e., how many distinct spin states and charge states are possible, respectively.

The \textbf{edge embedding} for atoms $i$ and $j$ is defined as
\begin{align}
A_{ij} = \psi_\text{edge} \Big( 
    &\left[ 
        \mathbf{W}^{\text{atom}} {h}(Z_i)
        + \mathbf{e}_{i}, \right. \notag\\
    &\left.\mathbf{W}^{\text{atom}} {h}(Z_j)
        + \mathbf{e}_{j} 
    \right] 
\Big) \ ,
\label{eqn:emb}
\end{align}
where 
$Z_i, Z_j \in \mathbb N$ are the atomic numbers and
$\mathbf{W}^{\text{atom}} \in \mathbb{R}^{D \times d_\text{atom}}$ are the learned embedding weights for the atomic numbers.
Lastly, $\psi_\text{edge}: \mathbb{R}^{2D} \rightarrow \mathbb{R}^{D}$ is an MLP that projects concatenated embeddings into the embedding space.

To embed the \textbf{pairwise distance} between atoms we use a set of $K=128$ radial basis functions (RBF), 
\begin{equation}
    \varphi_k(x)
    = \frac{1}{\sqrt{2\pi}\sigma_k}\exp\left(-\frac{(x - \mu_k)^2}{2\sigma_k^2}\right)
\end{equation}
with
learnable parameters $\mu_k$ and $\sigma_k$, where the former is initialized on $\mathcal{U}(0, 7)$ and the latter on $\mathcal{U}(0, 3)$,
to obtain
\begin{equation}
    R_{ij}
    = \psi_\text{dist}\left(\sum_{k=1}^K \varphi_k(m d_{ij} + b)\right),
\end{equation}
where $\psi_\text{dist}$ is an MLP and
$m$, $b$, are also learnable parameters initialized as 1 and 0, respectively
and $d_{ij}$ being the interatomic distance.

\textbf{Directional embeddings} are created by embedding the azimuthal and polar angles $\phi$ and $\theta$ of the pairwise displacement vectors using $K'=128$ Fourier kernels and a linear layer. The azimuthal frequencies and polar frequencies are defined as 
\begin{equation}
    \omega_\phi^{k'} = \pi \left(\frac{1}{2\pi}\right)^{\frac{k'}{K'/2-1}}
    \quad \text{and} \quad
    \omega_\theta^{k'} = \pi \left(\frac{1}{\pi}\right)^{\frac{k'}{K'/2-1}},
\end{equation}
which covers the angular space for a sufficient $K'$.
With the Fourier feature matrix
\begin{equation}
    F(\phi, \theta) 
    = \begin{bmatrix} 
        \sin(\phi\boldsymbol{\omega}_\phi) &
        \cos(\phi\boldsymbol{\omega}_\phi) &
        \sin(\theta\boldsymbol{\omega}_\theta) &
        \cos(\theta\boldsymbol{\omega}_\theta) 
    \end{bmatrix} 
\end{equation}
the directional embedding is 
\begin{equation}
    \Gamma_{ij} = \psi_\text{dir}\left(F(\phi, \theta)\right),
\end{equation}
where $\psi_\text{dir}$ is another linear layer.

Finally, all embeddings are added so the final edge embedding is
\begin{equation}
    E_{ij} = A_{ij} + R_{ij} + \Gamma_{ij}\,.
\end{equation}

\subsection{MD-ET implementation details}
\label{sec:et-impl-details}
\subsubsection*{Prediction head and training objective}
Forces are predicted directly, i.e., 
\begin{equation}
     {\boldsymbol{\hat f}}_i 
    = \psi_3
    \left(
        \sum^{N}_{l=1} 
            \psi_1(\boldsymbol{x}_{il}) + 
            \psi_2(\boldsymbol{x}_{lj})
    \right)
    \in \mathbb{R}^3,
\end{equation}
where $\psi_{\{1:3\}}$ are MLPs. 
The loss functional for training is
\begin{equation}
    \mathcal L
    = \frac{1}{N}\sum_{i=1}^{N} \|\boldsymbol{\hat f}_i - \boldsymbol{f}_i^*\|_2  \ ,
\end{equation}
 
where $\|\cdot , \cdot \|_2$ is the $\ell^2$ norm between prediction and the ground truth, ${\boldsymbol{f}}_i^*$, and $N$ is the number of atoms in the structure.

\subsubsection*{ET layer implementation details}
We implement MD-ET using PyTorch 2.5.1. Pseudo-code of the triangular attention mechanism and a comparison with regular (single-head) attention can be found in Algorithm 1 (below). To improve performance, we compile the triangular attention operation at runtime using \texttt{torch.compile}. This necessitates rewriting triangular attention without Einstein summation, \texttt{einsum}. 

\begin{algorithm}[ht]
\caption{Comparison of standard attention vs. triangular attention (TRIA)}
\begin{algorithmic}[1]
\centering
\begin{minipage}{0.45\textwidth}
\STATE \textbf{function} ATTENTION($\boldsymbol{X} : N \times D$)
\STATE \quad $\boldsymbol{Q}, \boldsymbol{K}, \boldsymbol{V} \leftarrow \text{linear}(\boldsymbol{X}).\text{chunk}(3)$
\STATE \quad
\STATE \quad $\tilde{\boldsymbol{A}} \leftarrow \text{einsum}(id, jd \rightarrow ij, \boldsymbol{Q}, \boldsymbol{K})$
\STATE \quad $\boldsymbol{A} \leftarrow \text{softmax}\left( \frac{\tilde{\boldsymbol{A}}}{\sqrt{d}} \right)$
\STATE \quad $\boldsymbol{O} \leftarrow \text{einsum}(ij, jd \rightarrow id, \boldsymbol{A}, \boldsymbol{V})$
\STATE \quad \textbf{return} $\text{linear}(\boldsymbol{O})$
\STATE \textbf{end function}
\end{minipage}
\hfill
\begin{minipage}{0.45\textwidth}
\STATE \textbf{function} TRIA($\boldsymbol{X} : N \times N \times D$)
\STATE \quad $\boldsymbol{Q}, \boldsymbol{K}, \boldsymbol{V}^1, \boldsymbol{V}^2 \leftarrow \text{linear}(\boldsymbol{X}).\text{chunk}(4)$
\STATE \quad $\boldsymbol{V} \leftarrow \boldsymbol{V}^1_{il} \odot \boldsymbol{V}^2_{lj}$  \# {Element-wise product}
\STATE \quad $\tilde{\boldsymbol{A}} \leftarrow \text{einsum}(ild, ljd \rightarrow ilj, \boldsymbol{Q}, \boldsymbol{K})$
\STATE \quad $\boldsymbol{A} \leftarrow \text{softmax}\left( \frac{\tilde{\boldsymbol{A}}}{\sqrt{d}} \right)$
\STATE \quad $\boldsymbol{O} \leftarrow \text{einsum}(ij, iljd \rightarrow ijd, \boldsymbol{A}, \boldsymbol{V})$
\STATE \quad \textbf{return} $\text{linear}(\boldsymbol{O})$
\STATE \textbf{end function}
\end{minipage}
\end{algorithmic}
\end{algorithm}

\newpage
\subsection{Inference speed evaluation}
\label{si:timings}
\begin{figure}[ht]
    \centering
    \includegraphics[width=0.85\linewidth]{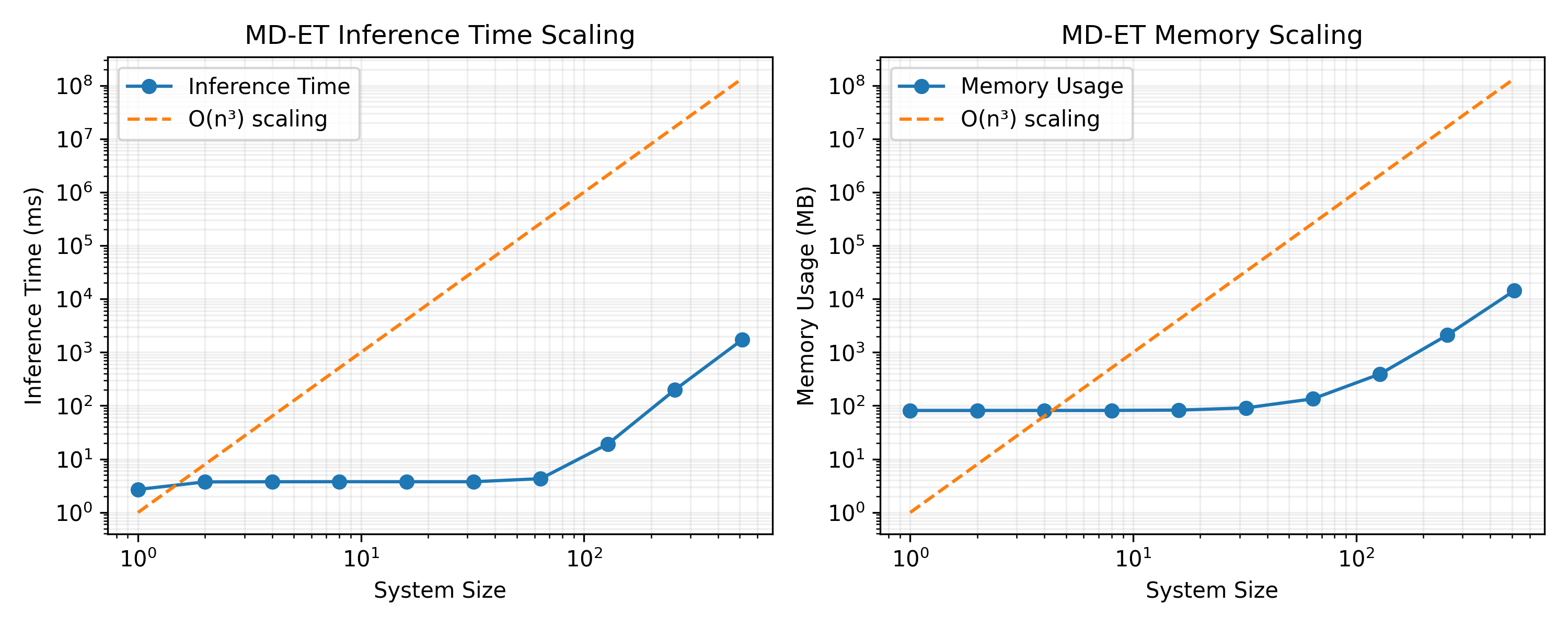}
    \caption{Empirical inference speed and memory footprint scaling of MD-ET.}
    \label{fig:scaling}
\end{figure}
Figure \ref{fig:scaling} shows empirical measurements of inference time and memory footprint scaling on an H100 PCIE GPU. We can see that MD-ET's cubic scaling becomes noticeable for system sizes larger than 100 atoms and problematic for systems larger than 500 atoms. See Table \ref{tab:speed} below for runtime benchmarks on MD17. Here, we time the forward pass of the model, including all postprocessing steps on a cloud-based V100 GPU (following the benchmark described in \cite{fu2022forces}). Inference Time measures the pure inference time without postprocessing.
\begin{table}[ht]
    \centering
    {\small$\begin{array}{ccccccc}
\toprule
\text{Molecule} & \text{Batch Size} & \text{Throughput} & \text{Avg Latency (ms)} & \text{Inference Time (ms)} & \text{FLOPs} \\
\midrule
\text{Aspirin} & 1 & 159.800 & 6.258 & 4.927 & 5.42 \times 10^9 \\
& 8 & 102.680 & 9.739 & 8.458 & 4.33 \times 10^{10} \\
& 64 & 18.685 & 53.520 & 52.103 & 3.47 \times 10^{11} \\
\hline
\text{Ethanol} & 1 & 148.151 & 6.750 & 5.429 & 9.96 \times 10^8 \\
& 8 & 153.357 & 6.521 & 5.247 & 7.97 \times 10^9 \\
& 64 & 70.741 & 14.136 & 12.784 & 6.37 \times 10^{10} \\
\hline
\text{Naphthalene} & 1 & 150.254 & 6.655 & 5.320 & 3.98 \times 10^9 \\
& 8 & 114.743 & 8.715 & 7.458 & 3.18 \times 10^{10} \\
& 64 & 22.033 & 45.387 & 43.976 & 2.55 \times 10^{11} \\
\hline
\text{Salicylic Acid} & 1 & 147.294 & 6.789 & 5.481 & 3.14 \times 10^9 \\
& 8 & 130.698 & 7.651 & 6.407 & 2.52 \times 10^{10} \\
& 64 & 27.834 & 35.927 & 34.564 & 2.01 \times 10^{11} \\
\bottomrule

\end{array}$
}
    \caption{Inference Speed Evaluation}
    \label{tab:speed}
\end{table}

\newpage
\subsection{Training protocol}
\label{si:training_protocol}
MD-ET was trained on the QCML dataset. We employed a cosine warmup learning rate schedule with an initial warmup phase of 5,000 steps, starting from 1e-6 and peaking at 5e-4. We conducted training on 8 NVIDIA A100 GPUs with 80 GB memory each. Important hyperparameters include:

\begin{table}[htbp]
    \centering
    \caption{Important pretraining parameters}
    \begin{tabular}{ll}
\toprule
\text{Parameter} & \text{Value} \\
\midrule
Learning Rate & $5 \times 10^{-4}$ \\
Batch Size & 1024 \\
Total Steps & 880,000 \\
Weight Decay & $1 \times 10^{-7}$ \\
Gradient Clip & 1.0 \\
Embedding Dimension & 192 \\
Number of Layers & 12 \\
Number of Heads & 12 \\
FFN Multiplier & 4 \\
3D Kernels & 128 \\
Warmup Steps & 5000 \\
Minimum Learning Rate & $5 \times 10^{-8}$ \\
Attention Dropout & 0 \\
FFN Dropout & 0 \\
\bottomrule
\end{tabular}    
    \label{tab:my_label}
\end{table}

\subsection{Experimental setup for benchmarks}
\label{si:experimental_setup_fine-tuning}

All fine-tuning experiments employ the AdamW optimizer with $\beta_1=0.9$, $\beta_2=0.999$, and weight decay $\lambda=1 \times 10^{-7}$. Experiments were conducted on a single NVIDIA H100 GPU with a batch size of 500. When the full batch exceeded GPU VRAM, gradient accumulation was used. The learning rate schedule combined cosine annealing with a linear warmup: fine-tuning lasted 2000 steps if not specified differently, with the first 100 steps as warmup, and a minimum learning rate of $1 \times 10^{-8}$. During fine-tuning, each sample was augmented with two transformations—random rotations and random reflections. 

\subsubsection{MD17-10k}
\label{sec:details-md17-10k}

For fine-tuning on the MD17 experiments we follow the experimental setup of \cite{fu2022forces}, namely we sample 9500/500/10000 conformations for train/val/test splits for each molecule and evaluate the force prediction error on the test set. The hyperparameters for each molecule are shown in Table \ref{tab:hyp_md17}.

\begin{table}[htbp]
  \centering
  \caption{Fine-tuning Hyperparameter for MD17}
  \label{tab:hyp_md17}
  \begin{tabular}{
    l
    c
    c
    c
    c
  }
    \toprule
    {Molecule} & {Train Split Size} & {Val Split Size} & {Test Split Size} & {Learning Rate} \\
    \midrule
    Aspirin          & 9500 & 500  & 10000 & $5 \times 10^{-5}$ \\
    Ethanol          & 9500 & 500  & 10000 & $5 \times 10^{-5}$ \\
    Naphthalene      & 9500 & 500  & 10000 & $5 \times 10^{-5}$ \\
    Salicylic Acid   & 9500 & 500  & 10000 & $5 \times 10^{-5}$ \\
    \bottomrule
  \end{tabular}
\end{table}

For full training runs on MD17 we use the same splits as above but train the model for 250k steps and with a batch size of 250, otherwise using identical settings to pretraining on QCML (see Appendix \ref{si:training_protocol})

For the stability evaluation we run an MD simulation with a timestep of 0.5 fs for 600k steps. We use a Nosé--Hoover thermostat set to 500K. For more details, see \citep{fu2022forces}.

\subsubsection{MD17@CCSD(T)}
\label{sec:details-md17-ccsd}

For the MD17@CCSD(T) experiments we adapt the experimental setup of \citet{chmiela2018towards} and extend the evaluation for MD-ET with a scenario with less training samples. Table \ref{tab:hyp_md17_ccsd} shows the hyperparameters for the MD17@CCSD(T) experiments.

\begin{table}[htbp]
  \centering
  \caption{Fine-tuning Hyperparameter for MD17@CCSD(T)}
  \label{tab:hyp_md17_ccsd}
  \begin{tabular}{
    l
    l
    c
    c
    c
    c
  }
    \toprule
    {Train Split Size} & {Molecule} & {Fine-tuning Steps} & {Test Split Size} & {Learning Rate} \\
    \midrule
    \multirow{5}{*}{1000 Samples} &
    Ethanol          & 2000  & 500 & $5 \times 10^{-5}$ \\
    & Aspirin          & 2000  & 500 & $5 \times 10^{-5}$ \\
    & Benzene      & 2000  & 500 & $5 \times 10^{-5}$ \\
    & Malonaldehyde   & 2000  & 500 & $5 \times 10^{-5}$ \\
    & Toluene   & 2000  & 500 & $5 \times 10^{-5}$ \\
    \midrule
    \multirow{5}{*}{100 Samples} &
    Ethanol          & 400  & 500 & $5 \times 10^{-5}$ \\
    & Aspirin          & 400  & 500 & $5 \times 10^{-5}$ \\
    & Benzene      & 400  & 500 & $5 \times 10^{-5}$ \\
    & Malonaldehyde   & 400  & 500 & $5 \times 10^{-5}$ \\
    & Toluene   & 400  & 500 & $5 \times 10^{-5}$ \\
    \bottomrule
  \end{tabular}
\end{table}

\subsubsection{Ko2020}
\label{sec:details-ko2020}
For the dataset introduced by \cite{ko2021} we follow the general experimental setup of Unke et. al. \cite{unke2021spookynet}. We additionally randomly split a validation set from the test split to optimize the learning rate. The hyperparameters used for fine-tuning for each system are shown in table \ref{tab:hyp_ko2020}.

\begin{table}[ht]
  \centering
  \caption{Fine-tuning Hyperparameter for Ko2020}
  \label{tab:hyp_ko2020}
  \begin{tabular}{
    l
    c
    c
    c
    c
  }
    \toprule
    {Molecule} & {Train Split Size} & {Val Split Size} & {Test Split Size} & {Learning Rate} \\
    \midrule
    C$_{10}$H$_2$/C$_{10}$H$_3^+$   & 9035 & 50  & 930 & $5 \times 10^{-4}$ \\
    Na$_{8}$/Cl$_8^+$               & 4493 & 50  & 480 & $1 \times 10^{-4}$ \\
    Ag$_3^{+/-}$                    & 9930 & 50  & 1030 & $5 \times 10^{-5}$ \\
    Au$_2$--MgO                     & 4468 & 50  & 480 & $5 \times 10^{-4}$ \\
    \bottomrule
  \end{tabular}
\end{table}

\subsubsection{xxMD}
\label{sec:details-xxmd}
For the experiments with the xxMD dataset we use the canonical splits of the xxMD-DFT subset provided by the authors. For fine-tuning experiments we use identical settings as for all other fine-tuning experiments, optimizing for 2000 steps. For training from scratch we use almost identical parameters to the QCML pretraining reported above, but reduce compute requirement by training for only 250,000 steps, using a batch size of 256 and a feed-forward layer multiplier of 2.  We only hyper-optimize the learning rate, optimal learning rates are reported in table \ref{tab:hyp_xxmd}. 
\begin{table}[htbp]
  \centering
  \caption{Fine-tuning Hyperparameters for xxMD}
  \label{tab:hyp_xxmd}
  \begin{tabular}{lcccc}
    \toprule
    {Subset} & {Learning Rate Direct Training} & {Learning Rate fine-tuning} \\
    \midrule
    Azobenzene      & $5 \times 10^{-4}$ & $5 \times 10^{-5}$ \\
    Dithiophene     & $5 \times 10^{-4}$ & $1 \times 10^{-4}$ \\
    Malonaldehyde   & $5 \times 10^{-4}$ & $1 \times 10^{-4}$ \\
    Stilbene        & $5 \times 10^{-4}$ & $1 \times 10^{-4}$ \\
    \bottomrule
  \end{tabular}
\end{table}

\subsubsection{SPICE}
\label{sec:details-spice}
We use the SPICEv1 dataset with a random 90\%/5\%/5\% split on the molecular level. Since our QCML-trained model is able to represent most elements up to Astatine, we do not filter based on molecular composition. We only filter conformers with an atomic force vector larger than 50 eV. See Table \ref{tab:spice_filtered} for details on filtered samples per subset. In total we use around 98\% of samples (compared to 85\% for EScAIP and MACE-OFF\cite{kovacs2023mace}). We fine-tune for 2000 steps and use the same hyperparameters as for other datasets above, only tuning the learning rate (see Table \ref{tab:hyp_spice}).
\begin{table}[htbp]
\centering
\caption{Filtered conformers from the SPICE dataset. Overall: 22900/1127651 samples filtered (2.03\%)}
\label{tab:spice_filtered}
\begin{tabular}{lrrrr}
\toprule
\multirow{2}{*}{Subset} & \multicolumn{2}{c}{Train+Val} & \multicolumn{2}{c}{Test} \\
\cmidrule(lr){2-3} \cmidrule(lr){4-5}
& Filtered/Total & \% & Filtered/Total & \% \\
\midrule
Pubchem & 21495/695212 & 3.09 & 1216/36644 & 3.32 \\
Monomers & 0/17700 & 0.00 & 0/900 & 0.00 \\
Dimers & 0/328020 & 0.00 & 0/14025 & 0.00 \\
Dipeptides & 187/32100 & 0.58 & 2/1750 & 0.11 \\
Solvated Amino Acids & 0/1200 & 0.00 & 0/100 & 0.00 \\
\midrule
\textbf{Total} & \textbf{21682/1074232} & \textbf{2.02} & \textbf{1218/53419} & \textbf{2.28} \\
\bottomrule
\end{tabular}
\end{table}
\begin{table}[htbp]
  \centering
  \caption{Fine-tuning Hyperparameters for SPICE}
  \label{tab:hyp_spice}
  \begin{tabular}{lcccc}
    \toprule
    {Subset} & {Train Split Size} & {Val Split Size} & {Test Split Size} & {Learning Rate} \\
    \midrule
    Pubchem          & 638,140 & 35,577  & 35,428 & $5 \times 10^{-4}$ \\
    Monomers          & 16,800 & 900  & 900 & $1 \times 10^{-4}$ \\
    Dimers      & 314,410 & 13,610  & 14,025 & $5 \times 10^{-4}$ \\
    Dipeptides   & 30,263 & 1,650  & 1,748 & $5 \times 10^{-4}$ \\
    Solvated Amino Acids & 1,150 & 50 & 100 & $1 \times 10^{-4}$ \\
    \bottomrule
  \end{tabular}
\end{table}

\subsection{Equivariance proof}
\label{si:equiv_proof}

See main text for the equivariance error, Eq. (\ref{eqn:eq_error}), which is 0 for an equivariant model. We first simplify the first term and substitute the equivariance property $f_\theta(\mathcal{R}\boldsymbol{x}) = \mathcal{R}f_\theta(\boldsymbol{x})$:  
\begin{equation}
   \mathbb{E}_{\mathcal{R}} 
   \left[ 
        \mathcal{R}^\top f_\theta(\mathcal{R}\boldsymbol{x}) 
    \right] 
   = \mathbb{E}_{\mathcal{R}} 
   \left[ 
        \mathcal{R}^\top \mathcal{R}f_\theta(\boldsymbol{x}) 
    \right]
    =
    \mathbb{E}_{\mathcal{R}} 
    \left[ 
        \boldsymbol{I}f_\theta(\boldsymbol{x}) 
    \right] 
    = \mathbb{E}_{\mathcal{R}} 
        \left[ 
            f_\theta(\boldsymbol{x}) 
        \right]
\end{equation}  
since $\mathcal{R}^\top \mathcal{R} = \boldsymbol{I}$ (identity matrix) for all $\mathcal{R} \in \text{SO}(3)$.
The expectation $\mathbb{E}_{\mathcal{R}}[f_\theta(\boldsymbol{x})]$ is over $\mathcal{R}$, but $f_\theta(\boldsymbol{x})$ does not depend on $\mathcal{R}$. So we obtain
\begin{equation}
   \mathbb{E}_{\mathcal{R}} 
        \left[ f_\theta(\boldsymbol{x}) \right] 
   = f_\theta(\boldsymbol{x}).
\end{equation}
For the 2nd term we also apply the equivariance property. We obtain
\begin{equation}
   \mathcal{S}^\top f_\theta(\mathcal{S}\boldsymbol{x}) = \mathcal{S}^\top \mathcal{S}f_\theta(\boldsymbol{x}) = \boldsymbol{I}f_\theta(\boldsymbol{x}) = f_\theta(\boldsymbol{x}).
\end{equation}
Therefore, the entire expectation over $x$ and $\mathcal{S}$ is
\begin{equation}
   \mathbb{E}_{x, \mathcal{S}} [||f_\theta(\boldsymbol{x}) - f_\theta(\boldsymbol{x})||] = 0.
\end{equation}
\qed
\newpage

\section{Uniform sampling of rotations via the 600-cell}
\label{app:600cell}
To obtain a set of 60 regularly spaced rotations in $\text{SO}(3)$ we use the 600-cell, a convex regular 4-dimensional polytope with 120 vertices. Since the 4 dimensional unit sphere $\mathbb{S}^4$ covers SO(3) exactly twice (double cover), $\mathbb{S}^3 / {\pm 1} \cong \text{SO}(3)$, the quaternion representations $(w, x, y, z)$ and its antipode $-(w,x,y,z)$ map to the same 3D rotation. We can thus use the coordinates of half of the 600-cell's 120 vertices to derive 60 uniformly spaced SO(3) rotations.

For the larger set of 360 uniform points, we combine the 60 points above with half of the 600-cell's 600 tetrahedral faces (each defined by four vertices spanning a 3-simplex) and compute their centroids. Under the double cover, antipodal centroids map to the same $\text{SO}(3)$ rotation, yielding 300 points.  
The union of these two subsets forms a 360-point configuration in $\text{SO}(3)$ and is uniform due to the 600-cell's symmetry.

\section{Comparison of 50 ns free energy surface on alanine dipeptide}
\label{app:alanine-fes}
\added{
To assess whether and how conformational distributions are impacted by the lack of energy conservation, we run a 50\,ns MD simulation of alanine dipeptide (more precisely \textit{N}-acetyl-L-alanine-\textit{N'}-methylamide) both with the baseline MD-ET model as well as an identical model using automatic differentiation to obtain energy-conserving force predictions. To obtain this model, we first train an MD-ET architecture with both a force and an energy prediction head for 250k steps. Note that in order to obtain stable second-order gradients we need to replace the directional embedding procedure based on azimuth and polar angles with an MLP, directly embedding normalized displacement vectors. This has only a minor impact on performance. After 250k steps of direct training we remove the force prediction head and replace the energy prediction head with one deriving energy-conserving forces using automatic differentiation. We then fine-tune this modified model for an additional 50,000 steps as described in \citet{bigi2024} and \cite{fu2025}. 
}

\begin{figure}[ht!]
    \centering
    \includegraphics[width=1.0\linewidth]{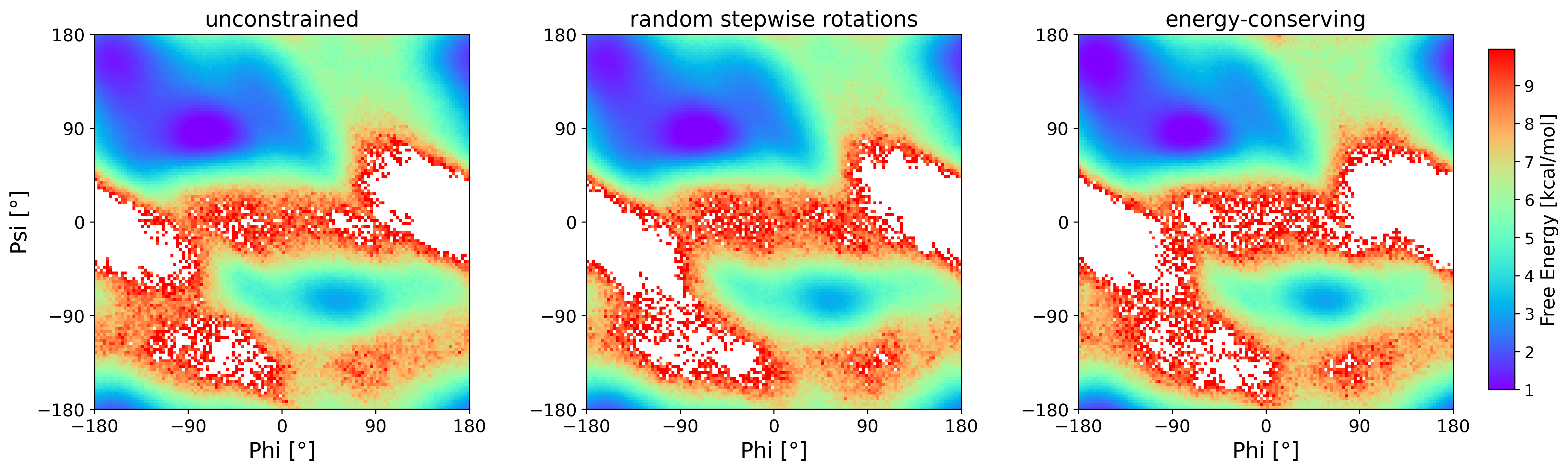}
    \caption{Comparison of 50 ns free energy surfaces of alanine dipeptide at 500 K using the SVR thermostat. Left: Baseline MD-ET model; middle: Baseline MD-ET model using random rotational offsets in each frame; right: Identical network architecture trained on identical data using energy-conserving forces.
    }
    \label{fig:alanine-fes-comparison}
\end{figure}

\added{
We run 50\,ns MD simulations with a step size of 0.5\,fs using the SVR thermostat set to a target temperature of 500\,K. We choose this temperature since the baseline model evaluated at fp32 precision shows a small but significant energy drift. To assess whether the random offset rotations described in Section \ref{sec:q3} impact observables, we run an additional 50\,ns simulation using the baseline model but with random offset rotations. The resulting free energy surfaces show no significant differences (see Figure \ref{fig:alanine-fes-comparison}).
}

\putbib[bib]
\defaultbibliographystyle{unsrtnat}
\end{bibunit}

\end{document}